\documentclass[10pt,twocolumn,letterpaper]{article}

\usepackage{cvpr}              % To produce the CAMERA-READY version
\usepackage{graphicx}
\usepackage{amsmath}
\usepackage{amssymb}
\usepackage{booktabs}
\usepackage{bbm}

\usepackage{enumitem}
\usepackage{amsmath}
\usepackage{mathtools}
\usepackage{amssymb}
\usepackage{amsthm, thmtools}

\usepackage{epsfig}
\usepackage{multirow}
\usepackage{makecell}
\usepackage{tabularx}
\usepackage{adjustbox}
\usepackage{booktabs}       % professional-quality tables
\usepackage{nicefrac}       % compact symbols for 1/2, etc.
\usepackage{algorithm} % For pseudo-code
\usepackage{algpseudocode} 
\usepackage{xspace}

\newcommand{\ours}{LENeRF\xspace}
\newcommand{\oursf}{Local Editing NeRF\xspace}
\newcommand{\lrm}{LRM\xspace}
\newcommand{\lrmf}{Latent Residual Mapper\xspace}
\newcommand{\afn}{AFN\xspace}
\newcommand{\afnf}{Attention Field Network\xspace}
\newcommand{\dn}{DN\xspace}
\newcommand{\dnf}{Deformation Network\xspace}

\usepackage[pagebackref,breaklinks,colorlinks]{hyperref}

\usepackage[capitalize]{cleveref}
\crefname{section}{Sec.}{Secs.}
\Crefname{section}{Section}{Sections}
\Crefname{table}{Table}{Tables}
\crefname{table}{Tab.}{Tabs.}

\def\cvprPaperID{5992} % *** Enter the CVPR Paper ID here
\def\confName{CVPR}
\def\confYear{2023}

\begin{document}

%%%%%%%%% TITLE - PLEASE UPDATE
\title{Local 3D Editing via 3D Distillation of CLIP Knowledge
}
% 후보: Local 3D Editing: NeRF and CLIP Complete Each Other
%NeRF+CLIP Knowledge Fusion \\
%LENeRF: Text-driven Local Editing of Geometry-aware 3D Generative Adversarial Networks
% LENeRF: Text-driven Local Editing of Geometry-aware 3D Generative Adversarial Networks

% =================== 저자 정보 (오리지널) =======================
%\author{Junha Hyung\\
%KAIST AI\\ 
%{\tt\small sharpeeee@kaist.ac.kr}
% For a paper whose authors are all at the same institution,
% omit the following lines up until the closing ``}''.
% Additional authors and addresses can be added with ``\and'',
% just like the second author.
% To save space, use either the email address or home page, not both
%\and
%Sungwon Hwang\\
%KAIST AI\\
%{\tt\small shwang.14@kaist.ac.kr}
%\and
%Daejin Kim\\
%Scatter Lab\\
%{\tt\small daejin@scatterlab.co.kr}
%\and
%Hyunji Lee\\
%KAIST AI\\
%{\tt\small alee6868@kaist.ac.kr}
%\and
%Jaegul Choo\\
%KAIST AI\\
%{\tt\small jchoo@kaist.ac.kr}
%}
%\maketitle

% =================== 저자 정보 (한줄짜리) (허용되는지 확인해보고 할 것) =======================

% \def\thefootnote{*}\footnotetext{Equal contribution}\def\thefootnote{\arabic{footnote}}

\newcommand*{\affaddr}[1]{#1}
\newcommand*{\affmark}[1][*]{\textsuperscript{#1}}
\newcommand*{\email}[1]{\texttt{#1}}
\author{
Junha Hyung\affmark[1,2]$^\dagger$\quad Sungwon Hwang\affmark[1]\quad Daejin Kim\affmark[3]\quad Hyunji Lee\affmark[1]\quad Jaegul Choo\affmark[1]\\
\\
\affaddr{\affmark[1]KAIST AI}\quad \affaddr{\affmark[2]Kakao Enterprise Corp.}\quad \affaddr{\affmark[3]Scatter Lab}\\
\small\email{\{sharpeeee, shwang.14, alee6868, jchoo\}@kaist.ac.kr, daejin@scatterlab.co.kr}
}

% \footnote{This work was done during an internship at Kakao Enterprise Corp.}

\twocolumn[{
\maketitle
\begin{center}
    \centering
    \captionsetup{type=figure}
    % \vspace{-2.0em}
    \includegraphics[width=1.0\linewidth]{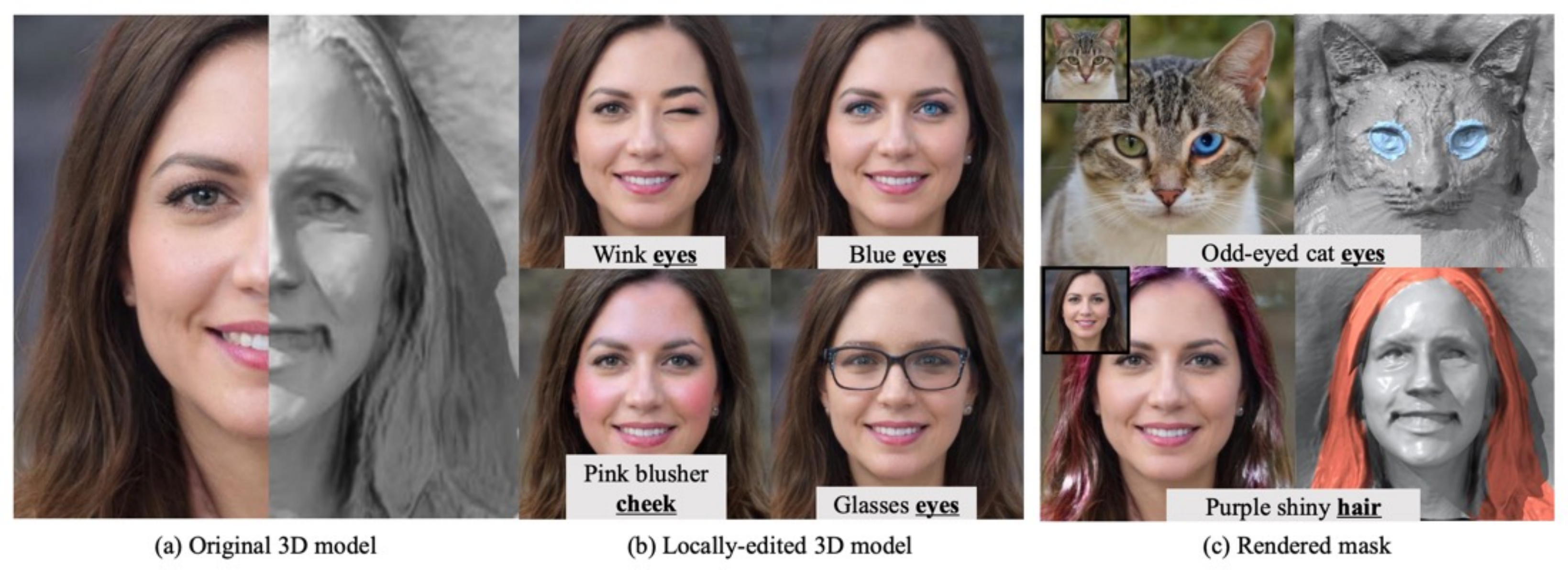}
    % \vspace{-0.2in}
    \captionof{figure}{Our local editing NeRF (LENeRF) enables users to edit specific areas of 3D assets based on textual prompts by estimating a 3D mask for tri-plane features. For instance, given an original 3D radiance field (a), users can define their desired area to edit (underlined text prompt, e.g., "eyes"). LENeRF then generates a 3D mask, which is employed for feature fusion, allowing for targeted modifications that adhere to the editing prompt (e.g., "blue eyes") (b). Additionally, as illustrated in (c), the 3D mask itself can be rendered and visualized for further analysis.}
    \label{fig: main}
% \vspace{-0.5em}
\end{center}
}]

% Section 제목이나 단락 사이 간격은 manually 정해져있어서, vspace 웬만하면 쓰지말것
% 페이퍼의 네모칸 (최소여백)을 해치는 vspace 사용하지말것!! (맨 위의 피겨인데 위의 공간을 또 줄인다거나)

\def\thefootnote{$\dagger$}\footnotetext{This work was done during an internship at Kakao Enterprise Corp.}\def\thefootnote{\arabic{footnote}}

\begin{abstract}

% 1108 SHWANG_EDITED
3D content manipulation is an important computer vision task with many real-world applications (e.g., product design, cartoon generation, and 3D Avatar editing). Recently proposed 3D GANs can generate diverse photorealistic 3D-aware contents using Neural Radiance fields~(NeRF). However, manipulation of NeRF still remains a challenging problem since the visual quality tends to degrade after manipulation and suboptimal control handles such as 2D semantic maps are used for manipulations. While text-guided manipulations have shown potential in 3D editing, such approaches often lack locality. To overcome these problems, we propose Local Editing NeRF~(LENeRF), which only requires text inputs for fine-grained and localized manipulation. Specifically, we present three add-on modules of LENeRF, the Latent Residual Mapper, the Attention Field Network, and the Deformation Network, which are jointly used for local manipulations of 3D features by estimating a 3D attention field. The 3D attention field is learned in an unsupervised way, by distilling the zero-shot mask generation capability of CLIP to the 3D space with multi-view guidance. We conduct diverse experiments and thorough evaluations both quantitatively and qualitatively.\footnote{We will make our code publicly available.}

\end{abstract}

\section{Introduction} 

\begin{figure}[t!]
\begin{center}
    \includegraphics[width=1.0\linewidth]{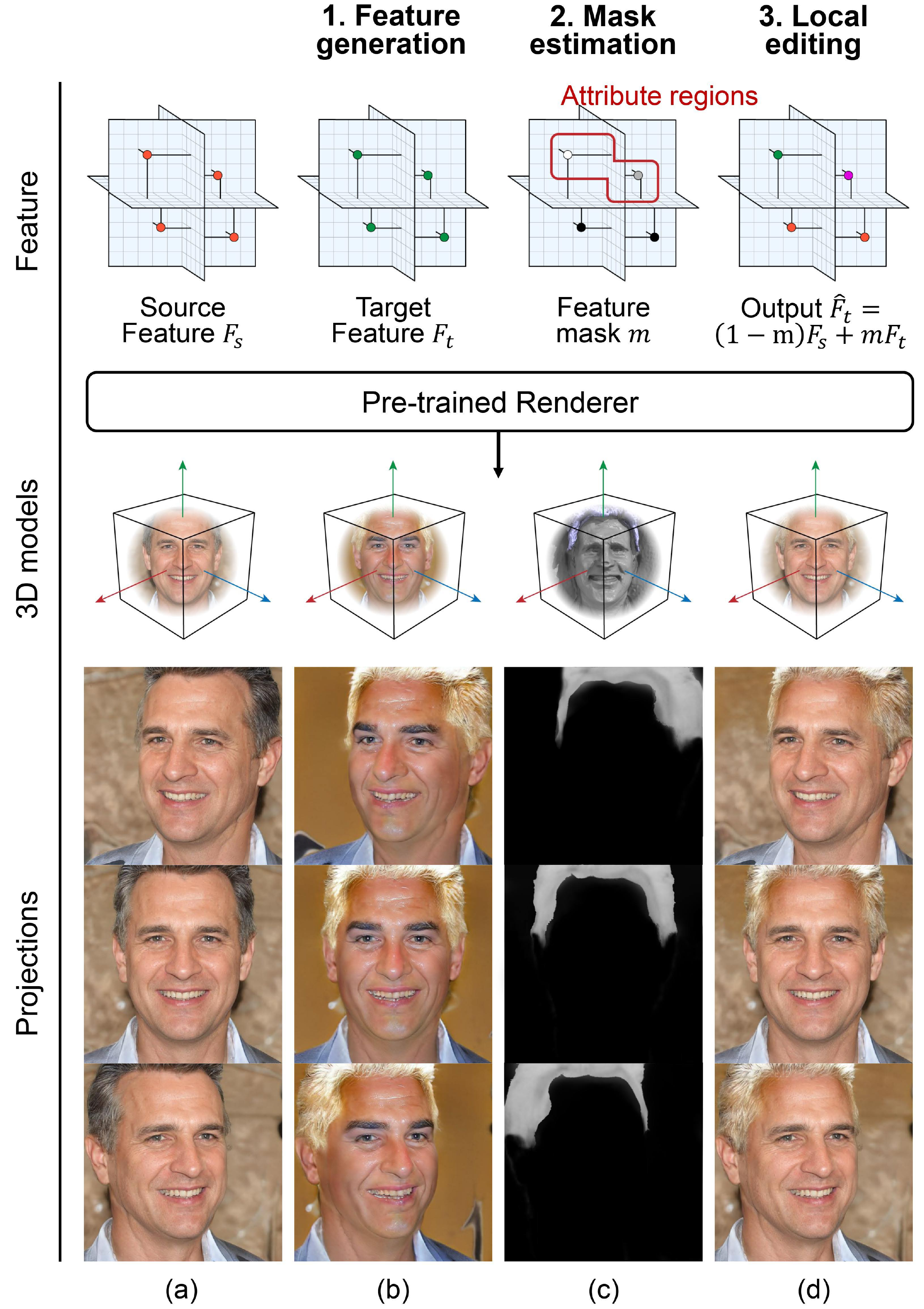}
\end{center}
    \vspace{-0.2in}
    \caption{Concept figure of \ours. Our method enables local editing of 3D assets by generating the target feature and estimating a 3D mask which guides the model on where to make changes at the feature level. Note that the mask is estimated for tri-plane features, not for raw RGB outputs.}
\label{fig_concepts}
\end{figure}

3D content editing has many real-world applications including but not limited to product design, cartoon generation, and 3D Avatar editing. However, it often necessitates the use of sophisticated tools with complex interfaces, which can be difficult for novice users and labor-intensive even for seasoned professionals.
%2 prev work 설명, limitations
%2-1 representing 3D
While explicit 3D representations such as voxels and meshes are commonly used for 3D generation and editing~\cite{DBLP:conf/iccv/GuillardRYF21, DBLP:conf/iccv/MehrJTCG19,DBLP:journals/tog/ZwickerPKG02}, they are memory-intensive and lack photorealism. In contrast, recent advances in Neural Radiance Fields (NeRF)~\cite{mildenhall2020nerf} have shown promising progress in representing 3D environments using implicit representations~\cite{DBLP:conf/cvpr/GenovaCSSF20,DBLP:conf/cvpr/JiangSMHNF20,DBLP:conf/cvpr/ParkFSNL19,mildenhall2020nerf} combined with volume rendering techniques that enable high-quality novel view synthesis. NeRF-based 3D GANs~\cite{DBLP:conf/nips/SchwarzLN020, DBLP:conf/cvpr/Niemeyer021,DBLP:conf/cvpr/ChanMK0W21,DBLP:conf/cvpr/DengYX022, DBLP:journals/corr/abs-2206-07255, DBLP:journals/corr/abs-2110-09788, DBLP:conf/iclr/GuL0T22,EG3D} have made further progress towards generating a category of 3D aware contents with a single model, extending the per-scene optimization scheme of NeRF. 

%2-2 previous works and limitations
Several studies~\cite{DBLP:conf/iccv/LiuZZ0ZR21, DBLP:conf/cvpr/WangCH0022, Sun_2022_CVPR, DBLP:journals/corr/abs-2205-15517} have attempted to address the challenges of NeRF editing, yet certain limitations persist. 
Works such as Edit-NeRF~\cite{DBLP:conf/iccv/LiuZZ0ZR21} and CLIP-NeRF~\cite{DBLP:conf/cvpr/WangCH0022} have pioneered NeRF manipulations, but they are constrained to low-resolution synthetic datasets and lack the capability to perform localized editing.
%Edit-NeRF~\cite{DBLP:conf/iccv/LiuZZ0ZR21} and CLIP-NeRF~\cite{DBLP:conf/cvpr/WangCH0022} take the first step in NeRF manipulations, but they operate only on a low-resolution synthetic dataset and cannot perform localized editing. 
Opposed to translation~\cite{DBLP:conf/cvpr/ChoiUYH20, DBLP:conf/iccv/ZhuPIE17} or style transfer~\cite{DBLP:conf/cvpr/GatysEB16} tasks, editing typically demands a certain degree of localization. 
However, achieving this with text-only control proves to be a challenging objective.
%ty, but is a difficult goal to achieve with text-only control. 
Alternative methods~\cite{Sun_2022_CVPR, DBLP:journals/corr/abs-2205-15517} that rely on semantic masks for editing face their own limitations: 2D guidance is not ideal for 3D editing and lacks the descriptiveness required for fine-grained editing. Furthermore, these approaches require inversion steps and are difficult to generalize across different domains, as they depend on the availability of labeled semantic masks.

%Other methods~\cite{Sun_2022_CVPR, DBLP:journals/corr/abs-2205-15517} rely on semantic masks for editing, but 2D guidance is suboptimal for 3D editing and is not descriptive enough to perform fine-grained editing. Also, they require inversion steps and are not easily generalizable to other domains since labeled semantic masks are necessary.

% 3 Our work
% 3-1 에디팅의 중요 요소
To overcome the existing limitations, we propose Local Editing NeRF (\ours), which focuses on the important aspects of 3D editing: photorealism, multi-view consistency, usability, diversity, and locality. With \ours, high-resolution photo-realistic radiance fields can be edited while maintaining their quality and multi-view consistency.
%Also, \ours is more usable in that it operates by \textit{text-only} editing and can be applied to any domain by utilizing the multi-modal embedding space of Contrastive Language Image Pre-training (CLIP)~\cite{CLIP}. Moreover, our method is capable of real-time editing as it does not have any test-time optimization process. 
One notable advantage of \ours is its text-only editing, making it more usable than other methods. This allows our approach to be applied to any domain by leveraging the multi-modal embedding space of Contrastive Language Image Pre-training (CLIP)~\cite{CLIP}. Additionally, our method achieves real-time editing as it does not require any test-time optimization process.

%4 구체적인 설명
Our proposed approach exhibits particularly robust performance in \textit{local} 3D editing.
This is achieved through a unique method of editing features in the 3D space independently by granting position-wise freedom to the features.
%\ours especially show robust performance on \textit{local} 3D editing by taking a whole new approach of editing the features in the 3D space independently by granting the features position-wise freedom. 
The naive approach of directly manipulating the latent code often results in \textit{global} changes to the 3D content, because features in the 3D space are spatially entangled with each other as the entire radiance field is conditioned with a single latent code.
To address this issue, we propose to generate a 3D mask on the region of interest with a masking prompt (e.g., "hair") and manipulate the features inside the region while leaving the rest unchanged.
Inspired by the previous approach which introduces the explanation method for capturing the regions of the interest~\cite{KimKC21}, we estimate 3D masks in an unsupervised fashion by using 3D distillation of the 2D CLIP model. Although the CLIP model is not 3D-aware and the 3D GAN lacks text-conditioned mask generation capability, our method enables the collaboration of two pre-trained models to generate a text-conditioned 3D mask, as demonstrated in Figure~\ref{fig: main} (c).
%complete each other to generate a text-conditioned 3D mask~(Figure~\ref{fig: main} (c)).

%%%%%%%%%%%
%Our proposed approach, Local Editing NeRF (\ours), exhibits particularly robust performance in local 3D editing. This is achieved through a unique method of editing features in the 3D space independently by granting position-wise freedom to the features. In contrast, the naive approach of directly manipulating the latent code often results in global changes to the 3D content. This is because all features in the 3D space are entangled with each other as the entire radiance field is conditioned with a single latent code.

%To address this issue, we propose generating a 3D mask on the region of interest using a location text prompt such as "hair." This allows us to manipulate the features inside the region while leaving the rest unchanged, resulting in more targeted and precise editing. We have found that 3D masks can be estimated in an unsupervised fashion using 3D distillation of the 2D CLIP model. Although the CLIP model is not 3D-aware and the 3D GAN lacks text-conditioned mask generation capability, two pre-trained models can work together to generate a text-conditioned 3D mask as shown in Figure~\ref{fig: main} (c).
%%%%%%%%%%%

\ours comprises three add-on modules, namely Latent Residual Mapper (\lrm), Attention Field Network (\afn), and Deformation Network (\dn) as depicted in Figure~\ref{fig_main}. \lrm generates a latent code that produces a target feature field. \afn generates a soft 3D mask indicating our region of interest. The source feature field is distorted using \dn and subsequently interpolated with the target field to synthesize the final feature field. 
\ours is trained with CLIP guidance~\cite{CLIP, Patashnik_2021_ICCV}, and \afn is additionally trained with CLIP-generated zero-shot pseudo labels.

The main contributions of our paper are as follows:
\begin{itemize}[noitemsep,leftmargin=1.8em]
    \item We introduce \oursf~(\ours), a 3D content editing framework capable of localized, photorealistic editing using a convenient real-time \textit{text-based} interface.

    \item Our method consists of add-on modules and does not require any domain-specific labels, allowing the method to be generalized to other models and domains.

    \item Our proposed technique involves a novel 3D distillation of CLIP knowledge, specifically an unsupervised approach that utilizes the 3D GAN and CLIP models jointly to generate 3D masks.
    
    \item We present diverse quantitative and qualitative results, along with various applications such as sequential editing, real image editing, and out-of-distribution editing.

\end{itemize}

\section{Related Work}
\label{gen_inst}

% \noindent\textbf{NeR}
% FENeRF: Face Editing in Neural Radiance Field	"1. Model is designed to render semantic mask out of shape code
% 2. Model is designed to render color out of texture code
% 3. Inverse render an image to retrieve shape and texture code
% 4. Replace semantic mask with desired semantic mask
% 5. Inverse render semantic mask to get edited shape code
% 6. Render with originnal texture code and inverse-rendered shape code to render edited image"
% CLIP-NeRF: Text-and-Image Driven Manipulation of Neural Radiance Fields	Text-and-Image Driven Manipulation of Neural Radiance Fields
% CoNeRF: Controllable Neural Radiance Fields	"Editing scene-nerf based on visual annotations
% (user scribble)"
% FEAT: Face Editing with Attention	CLIP-driven 2D StyleGAN editing with attention mask 
% Text and Image Guided 3D Avatar Generation and Manipulation	CLIP-driven 3D mesh editing (baseline: TBGAN)

\noindent\textbf{3D-Aware GANs} While Generative Adversarial Networks~(GANs)~\cite{GoodfellowPMXWOCB14} have demonstrated their ability to generate high-quality 2D images, generating 3D-aware content with GANs remains a challenging task. 
%Several approaches~\cite{DBLP:conf/iccv/Nguyen-PhuocLTR19, DBLP:conf/nips/Nguyen-PhuocRMY20} attempted to integrate GANs and neural scene representations, and 3D GANs~\cite{DBLP:conf/nips/SchwarzLN020, DBLP:conf/cvpr/Niemeyer021,DBLP:conf/cvpr/ChanMK0W21,DBLP:conf/cvpr/DengYX022, DBLP:journals/corr/abs-2206-07255, DBLP:journals/corr/abs-2110-09788, DBLP:conf/iclr/GuL0T22,EG3D, DBLP:conf/icml/KosiorekSZMSMR21} employing NeRF~\cite{mildenhall2020nerf} succeeded in generating high-quality images with multi-view consistency.
Several approaches have attempted to integrate GANs and 3D scene representations~\cite{DBLP:conf/iccv/Nguyen-PhuocLTR19, DBLP:conf/nips/Nguyen-PhuocRMY20}. Recent progress in 3D GANs~\cite{DBLP:conf/nips/SchwarzLN020, DBLP:conf/cvpr/Niemeyer021,DBLP:conf/cvpr/ChanMK0W21,DBLP:conf/cvpr/DengYX022, DBLP:journals/corr/abs-2206-07255, DBLP:journals/corr/abs-2110-09788, DBLP:conf/iclr/GuL0T22,EG3D, DBLP:conf/icml/KosiorekSZMSMR21} that employ NeRF~\cite{mildenhall2020nerf} have achieved success in generating high-quality, multi-view consistent images.
%pi-GAN~\cite{DBLP:conf/cvpr/ChanMK0W21} represents an implicit scene representation with periodic activation function for a view-consistent 3D generation. Meanwhile, EG3D~\cite{EG3D} leverages generated feature maps from StyleGAN2~\cite{Karras2019stylegan2} to construct a tri-plane feature field, which is volume rendered to conduct high-fidelity 3D aware image generations. 
%Our work is motivated by the advancements of the line of research, and we further enhance its applicability by suggesting a method to conduct text-guided local editing.
Building upon the progress made by previous research in this field, our work seeks to enhance its applicability by introducing a novel method for text-guided local editing.

%\paragraph{StyleGAN Manipulations} Many previous works have studied over 2D image manipulation tasks over StyleGAN \cite{DBLP:conf/cvpr/WuLS21, Patashnik_2021_ICCV, DBLP:journals/corr/abs-2104-08910, DBLP:conf/mm/YuZWZLCX0M22, DBLP:journals/corr/abs-2112-01573}. StyleSpace Analysis\cite{DBLP:conf/cvpr/WuLS21} studied the disentangling properties of the latent space of StyleGAN, followed by conducting attribute manipulations of generated images. StyleCLIP \cite{Patashnik_2021_ICCV}
%suggests several methods to map a given latent code to reflect the desired visual attributes of a text prompt. CF-CLIP\cite{DBLP:conf/mm/YuZWZLCX0M22} studies text-driven image manipulations toward attributes that are out of distribution from a dataset trained by a generator. Our work extends from these lines of research, as our editing method also focuses on manipulating the latent space of StyleGAN reflected in 3D representations.

\noindent\textbf{NeRF Manipulations}% EditNeRF
\ To enable local editing in 3D assets, EditNeRF~\cite{DBLP:conf/iccv/LiuZZ0ZR21} proposes a conditional NeRF that enables users to modify the color and shape of a specific region using scribbles. However, the editing ability of EditNeRF is limited to adding or removing local parts of 3D objects. 
Semantic mask guided 3D editing approaches~\cite{DBLP:conf/iccv/LiuZZ0ZR21,DBLP:conf/cvpr/KaniaYKTT22,DBLP:conf/cvpr/SunWZLZLW22,DBLP:journals/corr/abs-2205-15517} have made some progress in 3D editing. Among these works, FENeRF\cite{Sun_2022_CVPR} is trained to render 3D images and their corresponding semantic masks, allowing editing to be performed by inverting the model with edited semantic masks to modify 3D models.
%so that editing can be conducted by performing an inversion to the model with edited semantic masks to edit 3D models. 
%However, a 2D mask is a sub-optimal means for 3D editing, as texture or fine-grained attributes (e.g., expression) can not be manipulated with masks. Also, the methods require segmentation maps for both training and editing, which is not applicable to certain domains where segmentation maps are not retrievable.
However, using a 2D mask for 3D editing is suboptimal since fine-grained attributes such as expression or texture cannot be manipulated using masks.
Also, these methods require segmentation maps for both training and editing, which is not applicable to certain domains where segmentation maps are not available.
Meanwhile, there is a line of research utilizing CLIP~\cite{CLIP} for 3D generation and editing~\cite{DBLP:conf/cvpr/MichelBLBH22, DBLP:conf/cvpr/SanghiCLWCFM22, DBLP:conf/cvpr/JainMBAP22, DBLP:journals/tog/HongZPCYL22, DBLP:journals/corr/abs-2202-06079, DBLP:conf/cvpr/WangCH0022,DBLP:journals/corr/abs-2205-15585} that leverages the ease of manipulation using a text. However, the generated and edited content are not photorealistic and cannot provide fine-grained control. On the other hand, our work focuses on high-quality, localized manipulations that are well-suited for real-world applications.

\section{Preliminaries}
% \paragraph{NeRF}
\noindent\textbf{NeRF} NeRF~\cite{mildenhall2020nerf} is an implicit representation of 3D space using a MLP $\Psi$, as $(\textbf{c}, \sigma) = \Psi(\textbf{x}, \textbf{d})$, where $\textbf{x} = (x, y, z)$ is a point in 3D space, $\textbf{d}$ is a viewing direction, and resulting ($\textbf{c}, \sigma$) are color and volume density, respectively. The color and density values along the same ray are integrated to estimate a single pixel given camera extrinsic and pose as: 

\begin{equation} \label{eq:vr}
    \hat{C}(\textbf{r}) = \int_{k_n}^{k_f} T(k) \sigma(\textbf{r}(k)) c(\textbf{r}(k), \textbf{d}) \ dk,
\end{equation}

\noindent where $k_n$ and $k_f$ are near and far bounds, $\textbf{r}(k) = \textbf{o} + k\textbf{d}$ denotes a ray projected from the camera centered at position $\textbf{o}$, and $T(m) = \text{exp}(-\int_{k_n}^{m}\sigma(\textbf{r}(m))dm)$ is an accumulated transmittance along the ray. The parameters of $\Psi$ are optimized to produce $\hat{C}(\textbf{r})$, which are expected to be close to the ground truth pixel value.
%\vspace{-0.5em}

\noindent\textbf{StyleGAN Latent Code} The expressiveness of the StyleGAN latent code $w$ can be increased by inputting $k$ different latent codes, denoted by $\mathbf{w}$ ($\mathbf{w} \in \mathcal{W}^k \subsetneq \mathbb{R}^{k\times512}$)\footnote{For simplicity, we use the notations ($\mathcal{W}^k$, $\mathcal{W}^{k}_{*}$) from~\cite{tov2021designing}.}, into each of the $k$ StyleGAN layers.
%$\mathbf{w}$ to each $k$ StyleGAN layers ($\mathbf{w} \in \mathcal{W}^k \subsetneq \mathbb{R}^{k\times512}$)\footnote{For simplicity, we use the notations ($\mathcal{W}^k$, $\mathcal{W}^{k}_{*}$) from~\cite{tov2021designing}.}.
%inputting $k$ different latent codes $\mathbf{w} \in \mathcal{W}^k \subsetneq \mathbb{R}^{k\times512}$ to each of $k$ StyleGAN layers. 
This can further be extended to $\mathcal{W}^{k}_{*}$ as introduced in \cite{tov2021designing}.
Unlike $\mathcal{W}^{k}$, whose distribution is bounded by the mapping network, $\mathcal{W}^{k}_{*}$ can reside outside the range of the mapping network.
%it which can reside outside the range of a mapping network unlike $\mathcal{W}^{k}$ whose distribution is bounded by the mapping network.
%where unlike $\mathcal{W}^{k}$ whose distribution is bounded by a mapping network, distribution of $\mathcal{W}^{k}_{*}$ can reside outside of the range of mapping network. 
We choose to utilize $\mathcal{W}^{k}_{*}$ in our pipeline to enable a wider range of attribute manipulation. 

%denoted as $\pi_{vol}$
\noindent\textbf{EG3D} EG3D~\cite{EG3D} is a 3D-aware GAN based on NeRF. Feature vectors of a radiance field are sampled from tri-plane generated by StyleGAN2~\cite{Karras2019stylegan2} and summed to a tri-plane feature vector $F = g_{\theta}(\mathbf{x}, \mathbf{w})$ given a 3D position $\mathbf{x} \in \mathbb{R}^3$ and a latent code $\mathbf{w}$ that modulates the StyleGAN2 layers, and $g_{\theta}$ denotes the function of the whole process.
$F$ is further processed by a small decoder to a $M_f$-dimensional feature vector. 
Then a set of sampled features $\{{F_{ij}}\}_{j=1}^{N_s}$ along the set of camera rays $\{\mathbf{r}_i\}_{i=1}^{H_{V}\times W_{V}}$ is aggregated with volumetric rendering using Eq.~\ref{eq:vr}, where $N_s$ is the number of sampled points per ray, and $H_{V}$ and $W_{V}$ are the height and the width of the rendered feature image, respectively.
Finally, the feature image is decoded into the high-resolution image by a super-resolution module $\pi^{SR}$:

\begin{equation} \label{eq:sr}
\begin{aligned}
 \pi^{SR}: \mathbb{R}^{H_V \times W_V \times M_f} \rightarrow \mathbb{R}^{H \times W \times 3}
\end{aligned}
\end{equation}

\begin{figure*}[t!]
\begin{center}
    \includegraphics[width=1.0\linewidth]{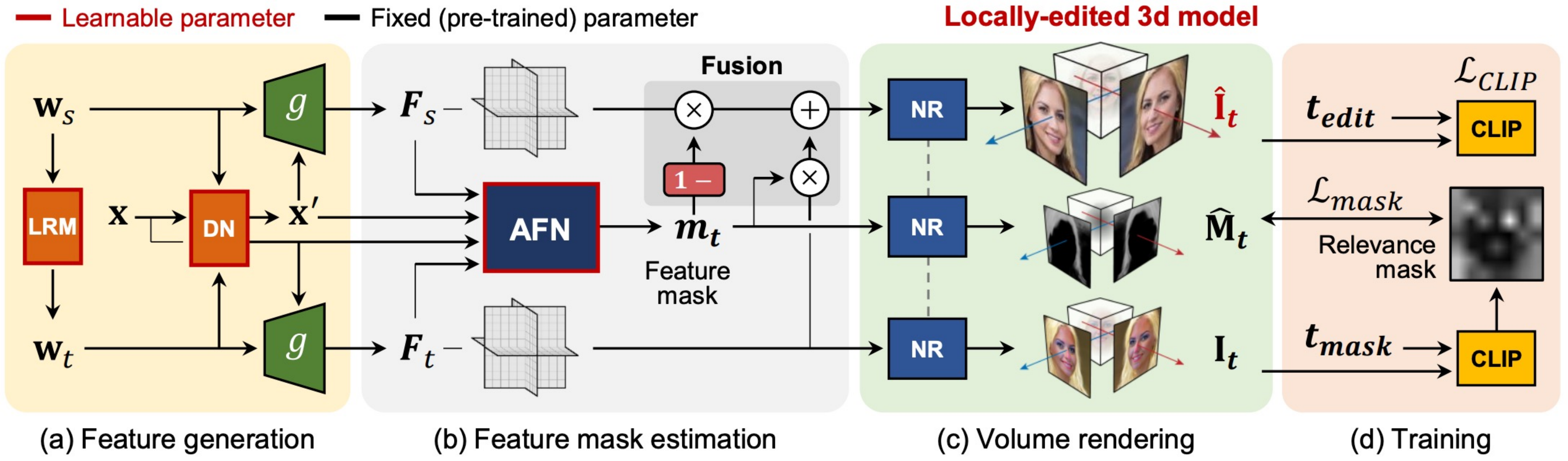}
\end{center}
    \vspace{-0.2in}
    \caption{The overall architecture and the training objective of LeNeRF. (a) The target latent code and the deformation field are generated by the latent residual mapper (LRM) and the deformation network (DN), respectively. (b) The tri-plane features for the source and target are encoded by the pretrained generator and the soft mask for combining both features is estimated by our attention field network (AFN). (c) By using the neural renderer (NR) which incorporates the volume rendering and a super-resolution (not applied to rendered masks), we obtain the rendered results for the source and target features, mask, and the locally-edited features, which is the final output of LeNeRF. (d) In the training process, only LRM, DN, and AFN are trained for local editing, while other parameters for the generator and neural renderer are fixed.}
\label{fig_main}
\end{figure*}

\section{LeNeRF Framework for 3D Local Editing}
In this section, we present \oursf (\ours) which enables \textit{localized} editing of 3D scenes using a text prompt for manipulation, $t_{edit}$ (e.g., \textit{blue eyes}), and a prompt specifying the region of interest, $t_{mask}$ (e.g., \textit{eyes}). 
Unlike previous methods that perform global editing by updating the latent codes of 3D GAN~\cite{DBLP:conf/cvpr/WangCH0022}, we propose generating a 3D mask using CLIP~\cite{CLIP} supervision and performing localized editing with the generated mask through a \textit{position-wise} transformation of feature fields.
%Instead of performing global editing by updating the latent codes of 3D GAN as in previous work~\cite{DBLP:conf/cvpr/WangCH0022}, we propose to generate 3D mask only with the CLIP~\cite{CLIP} supervision and perform local editing with the generated mask via \textit{position-wise} transformation of feature fields.

As shown in Fig.~\ref{fig_main}, \ours consists of frozen pretrained NeRF generator~\cite{EG3D} along with trainable add-on modules: \lrmf (Section~\ref{method:lrm}), \afnf (Section~\ref{method:rff}), and \dnf (Section~\ref{method:hld}). 
%The Latent Residual Mapper outputs latent codes $\mathbf{w}_t \in \mathcal{W}^k_*$ which generates intermediate feature fields, given the source latent codes $\mathbf{w}_s \in \mathcal{W}^k$. 
The \lrmf (\lrm) maps the source latent code $\mathbf{w}_s \in \mathcal{W}^k$ to the target latent code $\mathbf{w}_t \in \mathcal{W}^k_*$, which in turn conditions the source and target feature vectors $F_s$ and $F_t$ respectively (Fig.~\ref{fig_main} (a)).

Then the 3D soft mask corresponding to the regions specified by $t_{mask}$ is estimated by the \afnf (\afn) (Fig.~\ref{fig_main} (b)). The Deformation Network handles geometric manipulations by deforming the source feature field $\{F_s\}$, and the source and target features are fused to produce the final feature fields. The modules are trained using the proposed CLIP loss $\mathcal{L}_{CLIP^{+}}$, and the AFN is trained additionally with the mask loss $\mathcal{L}_{mask}$ via pseudo-labels generated with relevance map aggregation conditioned on the mask prompt $t_{mask}$ (Fig.~\ref{fig_main} (d), Section~\ref{method:training}). Once trained, LeNeRF performs 3D editing in real-time.

%by using radiance

%Plausible target features that is g

% LENeRF fuse the feature field of source image ($\mathcal{R}_s$) with manipulated feature field ($\mathcal{R}_t$) using 3D attention field. Our method utilizes frozen pretrained NeRF generator~\cite{corr/abs-2112-07945} along with trainable add-on modules: Latent Residual Mapper (Section~\ref{method:lrm}), Attention Field Network (Section~\ref{method:rff}), and Deformation Network (Section~\ref{method:hld}). Once trained, manipulation is performed at an interactive rate. Please see the supplement for more details.
 %(Section~\ref{sec:training}). 

\subsection{\lrmf (\lrm)} \label{method:lrm}
% 1. mapper에 대한 설명 (Figure 1의 어느 부분이 mapper다)
% 2. 차별점
% 3. 수식

% mapper를 왜 쓰는지 어떤 건지 
Inspired by CLIP-guided 2D image editing techniques~\cite{Patashnik_2021_ICCV, DBLP:journals/tog/GalPMBCC22, DBLP:conf/mm/YuZWZLCX0M22, DBLP:journals/corr/abs-2104-08910, DBLP:conf/siggraph/AbdalZ0MW22}, we train the \lrmf~(Figure~\ref{fig_main} (a)), a mapper function that generates the target latent code $\mathbf{w}_t$ given the source latent code $\mathbf{w}_s$. 
The mapper is trained using CLIP guidance, and once trained, manipulation can be performed in real-time without the need for any inversion steps~\cite{Patashnik_2021_ICCV}.

\begin{figure*}[t!]
\centering
\includegraphics[width=0.9\textwidth]{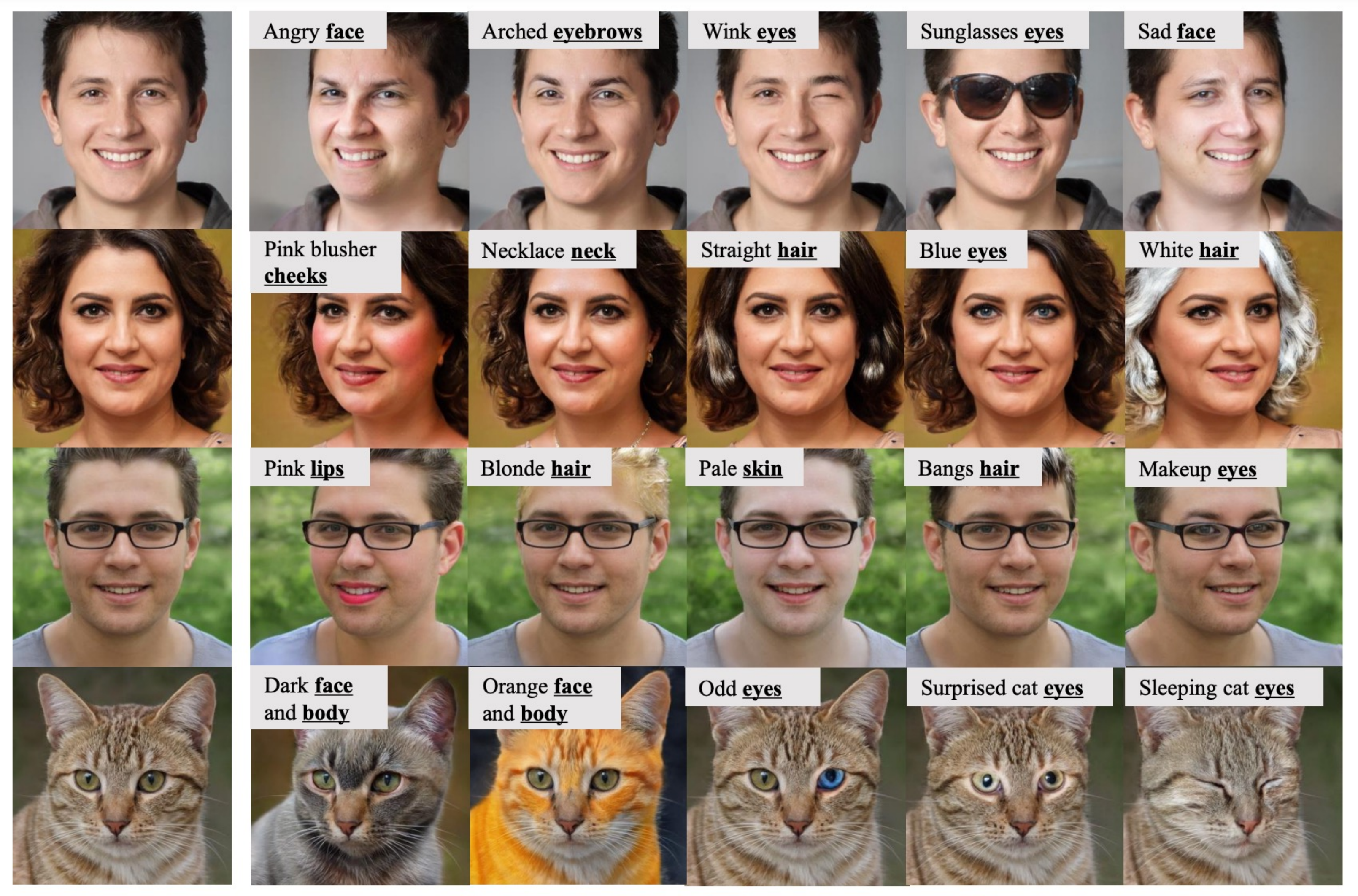}
%\vspace{-0.2in}
\caption{Curated Examples of \ours.}
\label{fig:curated}
\vspace{-1.0em}
\end{figure*}

Instead of using $\mathbf{w}_t$ directly for generating the final radiance field, we utilize the mapper to produce tri-plane features $\{F_{t}\}$, which are composited with the source tri-plane feature field $\{F_{s}\}$ to generate the final feature field. This composition involves interpolating two features based on an estimated 3D mask~(Section~\ref{method:rff}).

The features in the 3D space are spatially entangled because the entire feature field is conditioned with a single latent code. By utilizing more than one latent code (in our case, two), we can grant position-wise freedom to the features, which enables the localized editing of the source features based on the 3D mask and the target features. Specifically, given the source latent code $\mathbf{w}_s \in \mathcal{W}^k$, the mapper outputs the target latent code $\mathbf{w}_t \in \mathcal{W}^k_*$ which is used to produce the final feature field, described as:

\begin{equation}
\begin{aligned}
M(\mathbf{w}_s) &= (M^{1}(\mathbf{w}_s^1), ... , M^{N}(\mathbf{w}_s^N)) \\
&= (\Delta\mathbf{w}^1, ... , \Delta\mathbf{w}^N), \\
\end{aligned}
\end{equation}
\begin{equation}
\begin{aligned}
\mathbf{w}_{i} = (\mathbf{w}_s^1+\Delta\mathbf{w}^1, ..., \mathbf{w}_s^N+\Delta\mathbf{w}^N),
\end{aligned}
\end{equation}
where $M^{i}, i \in {1, 2, ..., N}$ are fully-connected networks and $\mathbf{w}^{n}$ denotes the subset of the latent code $\mathbf{w}$ that modulates the $n$-th group of StyleGAN2-like backbone layers. 

We design our mapper to estimate the residual values of latent codes, which enables easier regularization and training. \lrm is trained to minimize the $\mathcal{L}_{CLIP^+}$ objective given a target prompt $t_{edit}$. See Section~\ref{method:training} for more details.

\subsection{Attention Field Network (\afn)} \label{method:rff}

%3D attention field - module 이름 attention field network (AFN)
%(이 부분 distentangle 에 대한 부분,, AFN의 당위성을 설명하는데 필요하지 않나??)
%While latent residual mapper alone can perform radiance field manipulation to some extent, we find it extremely difficult to perform fine-grained and localized editing. We find that there are two reasons for this: First, manipulation of $\mathbf{w}_s$ has global impact on the radiance field, and does not guarantee local changes. We show this by performing 3D ediitng via only using latent residual mapper. First \textcolor{red}{result(Fig)} indeed show that even with careful tuning of networks and losses, different attributes such as expression, skin color, and background are tied together. Next, we discover that latent space of generative models in general are highly biased towards data distribution, therefore failing to produce uncommon results even with proper guidance. For example, bald people are mostly middle-aged male in FFHQ~\cite{DBLP:conf/cvpr/KarrasLA19} dataset, therefore such manipulations generally corrupt the results, or lead to the shifting of identities to middle-aged male. \textcolor{red}{result(Fig)} 
The Attention Field Network~(AFN) parameterizes the 3D attention field that serves as a soft 3D mask for interpolation between the source and the target features. It is conditioned on both the source and target features, 3D position, and the corresponding latent codes, allowing the model to take into account both point semantics and global context:
%which enables the model to be informed of both the point semantics and the global context:

%Therefore, in order to perform fine-grained and localized 3D editing, we build our model to control and manipulate features in 3D space locally, conditioned by source features and their positions. We first estimate 3D attention field that decides the rate of interpolation between two radiance(feature) fields $\mathcal{R}_s$ and $\mathcal{R}_t$ conditioned by $\mathbf{w}_s$ and $\mathbf{w}_t$. Then we fuse two fields using the estimated values, which enables local manipulations of our region of interest. This not only enables local editing, but also collage of radiance fields enable counterfactual or out-of-distribution manipulations without destroying the original identities (e.g., bald person).

% 이부분은 suppl로?
%is highly entangled. We manipulate each dimension of EG3D latent code $w \in \mathcal{W}$ and observe $G(w)$. Fig shows that each dimension of $w$ effects many attributes at once(e.g. background, gender, skin color) and mostly has global impact on the image rather than changing local semantics. While EG3D utilizes StyleGAN-like backbone, mechanism of $w$ differs from that of 2D StyleGAN because it generates 3D features from tri-plane, which represents radiance field features, while $w$ in 2D StyleGAN represents 2D pixel features

% m(\textbf{x}) = A_{\theta_{a}}(\textbf{f}_{s}(\textbf{x})\oplus\textbf{x},\textbf{f}_{i}(\textbf{x})\oplus \textbf{x},\mathbf{w}_{s}, \mathbf{w}_{i}),
% 이거 centering 어케함
\begin{equation} \label{eq:mx}
\begin{aligned}
A:  \mathbb{R}^{2M_{f}+6} \times \mathbb{R}^{2d} \rightarrow [0,1], \\
m = A(F_{s}\oplus \mathbf{x},F_{t}\oplus \mathbf{x},\mathbf{w}_{s}, \mathbf{w}_{t}),
\end{aligned}
\end{equation}
\begin{equation} \label{eq:fusion}
\begin{aligned}
\hat{F} = (1-m)*F_s + m*F_{t},
\end{aligned}
\end{equation}
where $F_{s}$, $F_{t} \in \mathbb{R}^{M_f}$ are the source and target tri-plane feature vectors, each conditioned by $\mathbf{w}_{s}, \mathbf{w}_{t}  \in \mathbb{R}^d$ respectively, and $m \in \mathbb{R}$ is the value of soft mask at $\mathbf{x}$.

%Generated 3D attention field is then used to composite source and intermediate feature fields $\mathbf{f}_{s}$ and $\mathbf{f}_{t}$ and produce final feature field $\hat{\mathbf{f}}$ as:

For a position $\mathbf{x}$ inside the region of interest described by $t_{mask}$, AFN is trained to produce a high mask value $m$ so that the estimated feature $\hat{F}$ is dominated by the target feature $F_{t}$. For a position $\mathbf{x}$ outside of our region of interest, low mask values are estimated to ensure that the source feature remains unchanged.
Utilizing an estimated 3D soft mask enables 3D editing focused on desired regions, while avoiding unintended changes in other areas. Note that we do not use any domain-specific mask prediction model or semantic labels for AFN training. Instead, we show that it is possible to distill CLIP's capability to generate zero-shot 2D masks into 3D using our proposed loss function, as explained in Section~\ref{method:training}. 

%input이 저렇게 들어가는 이유
%network 구조

\subsection{Handling Large Deformations} \label{method:hld}
Deformations are necessary for manipulations that involve large geometric changes (e.g., opening mouth), as interpolating between features representing different semantics (e.g., teeth and lips) often leads to artifacts (please refer to the supplement for figures). The deformation field for the source radiance field is estimated by the Deformation Network $T$ conditioned with latent codes ($\mathbf{x}^{'} = T(\mathbf{x}, \mathbf{w}_{s}, \mathbf{w}_{t}) + \mathbf{x}$),
% \begin{equation}
% \begin{aligned}
% \mathbf{x}^{'} = T(\mathbf{x}, \mathbf{w}_{s}, \mathbf{w}_{t}) + \mathbf{x},
% \end{aligned}
% \end{equation}
and Eq. \ref{eq:mx} can be re-written as:
\begin{equation}
\begin{aligned} \label{eq:final_feature}
m = A(F_{s}\oplus \mathbf{x}^{'},F_{t}\oplus \mathbf{x},\mathbf{w}_{s}, \mathbf{w}_{t}).
\end{aligned}
\end{equation}

\subsection{Training} \label{method:training}
\noindent\textbf{Generator} The estimated features (Eq.~\ref{eq:final_feature}) are decoded into a radiance field and rendered into a 2D image $\hat{I}_t$ via volume rendering (Eq.~\ref{eq:vr}) followed by the super-resolution module $\pi^{SR}$ (Eq.~\ref{eq:sr}) given a randomly sampled camera pose $\mathbf{v} \sim \mathcal{Z}_v$, where $\mathcal{Z}_v$ is the camera pose distribution. Similarly, the source image $I_s$ and the raw target image $I_t$ are rendered using the same process with $F_s$ and $F_t$.
\\

\begin{figure}[t!]
\centering
\includegraphics[width=0.9\linewidth]{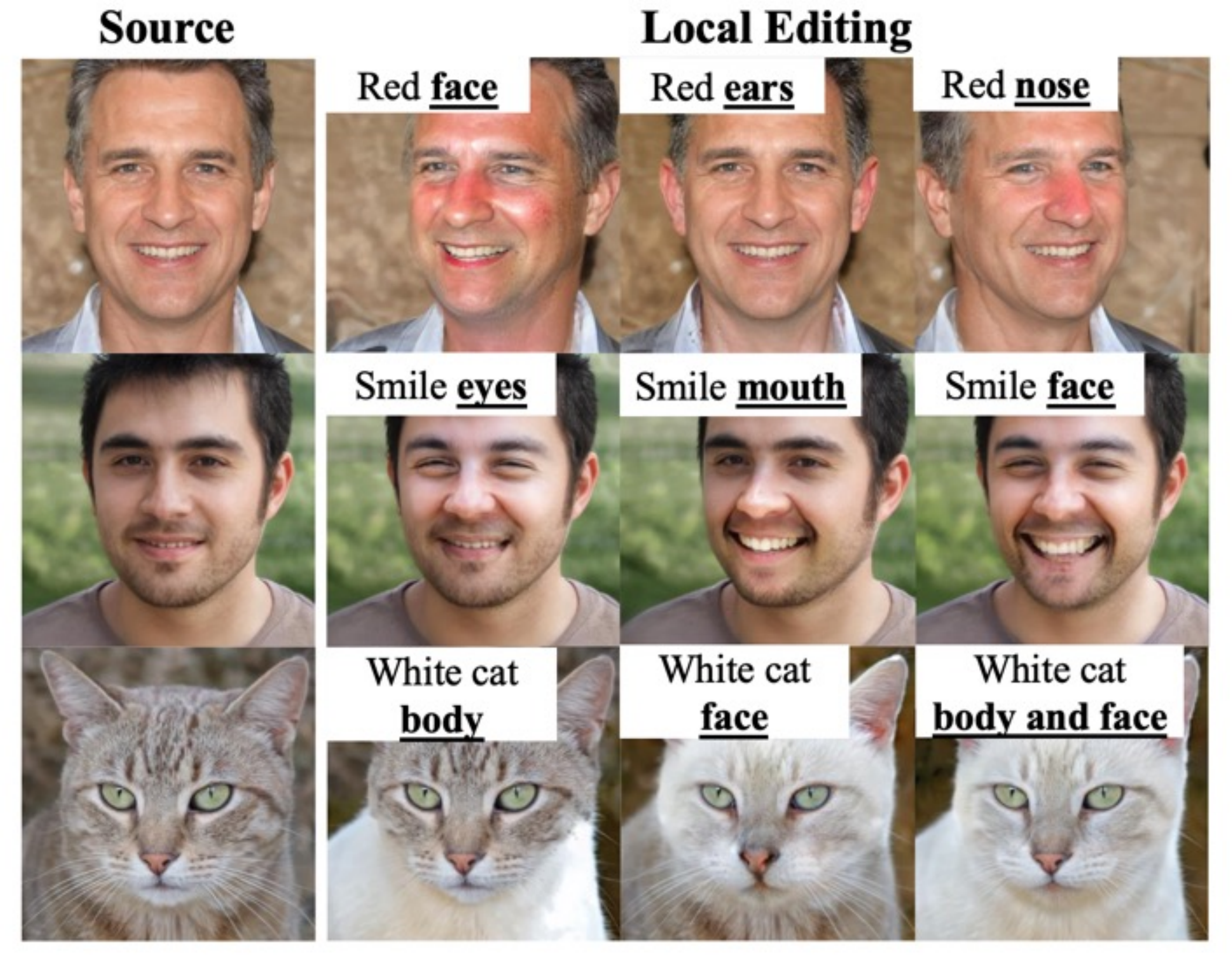}
%\vspace{-0.5em}
\caption{Results of partial editing.}
\label{fig: partial_editing}
%\vspace{-1.0em}
\end{figure}

\noindent\textbf{CLIP Loss} 
We use pretrained CLIP~\cite{CLIP} text encoder $E_{T}$ and image encoder $E_{I}$ to optimize our modules so that the estimated image $\hat{I}_t$ matches $t_{edit}$. However, a direct approach to maximize the cosine similarity of $E_{I}(\hat{I}_t)$ and $E_{T}(t_{edit}))$ frequently leads to degenerate solutions. This is because the CLIP embedding space includes false positives, where degenerate images result in a high similarity score with the target text embedding. Therefore, we instead propose a robust $\mathcal{L}_{CLIP}$ loss that utilizes contrastive loss together with image augmentations:
%$\hat{I} = \pi^{SR}_{\phi}(\pi_{vol} (\hat{\mathcal{R}}))$
\begin{equation}
\begin{aligned}
\mathcal{L}_{CLIP} = -\log{\frac{\sum_{s^{+}\in S^{+}} e^{q\cdot s^{+}}}{\sum_{s \in S^{+} \cup S} e^{q \cdot s}}},
\end{aligned}
\end{equation}
where $S^{+}$ is a set of positive samples\footnote{$S^{+} \vcentcolon= \{E_{T}(t_{edit}) - E_{I}(aug(I_{s})), E_{T}(t_{edit}) - E_{T}(t_{src})\}$} and $S^{-}$ is a set of negative samples\footnote{$S^{-} \vcentcolon= \{E_{T}(t_{edit}) - E_{I}(aug(I_{s}))\}$}.
% \begin{equation}
% \begin{aligned}
% &S^{+} \vcentcolon= \{E_{T}(t_{edit}) - E_{I}(aug(I_{s})), E_{T}(t_{edit}) - E_{T}(t_{src})\} \\
% &S^{-} \vcentcolon= \{E_{T}(t_{edit}) - E_{I}(aug(I_{s}))\} .
% \end{aligned}
% \end{equation}
Here, $q = E_{I}(aug(\hat{I}_t)) - E_{I}(aug(I_{s}))$, $t_{src}$ is a neutral text describing the source image (e.g., "Photo of a person"), and $aug$ refers to a random image augmentation. $\mathcal{L}_{CLIP}$ maximizes the mutual information between $q$ and the positive samples while minimizing that of the negative samples. Comparing the directions of the CLIP embeddings and using multiple cross-domain positive samples lead to stable and robust optimizations, along with the negative samples inhibiting lazy manipulations. 

Also, like ~\cite{Patashnik_2021_ICCV}, we incorporate an identity preserving loss~\cite{DBLP:conf/cvpr/RichardsonAPNAS21} $\mathcal{L}_{id}$, using a pretrained ArcFace~\cite{DBLP:journals/pami/DengGYXKZ22} model, and a $L_2$ norm of the manipulation step $M(\mathbf{w}_s)$ to preserve the attributes of the source radiance fields. The total loss for training the Latent Residual Mapper and the Deformation Network is $\mathcal{L}_{CLIP^{+}} = {\mathcal{L}_{CLIP} + \lambda_{L2} \lVert M_(\mathbf{w}_s) \rVert}_2 + \lambda_{id}\mathcal{L}_{id}$.
\vspace{0.5em}
% \begin{equation}
% \begin{aligned}
% \mathcal{L}_{CLIP^{+}} = {\mathcal{L}_{CLIP} + \lambda_{L2} \lVert M_(\mathbf{w}_s) \rVert}_2 + \lambda_{id}\mathcal{L}_{id}.
% \end{aligned}
% \end{equation}

\begin{figure}[t!]
\centering
\includegraphics[width=1.0\linewidth]{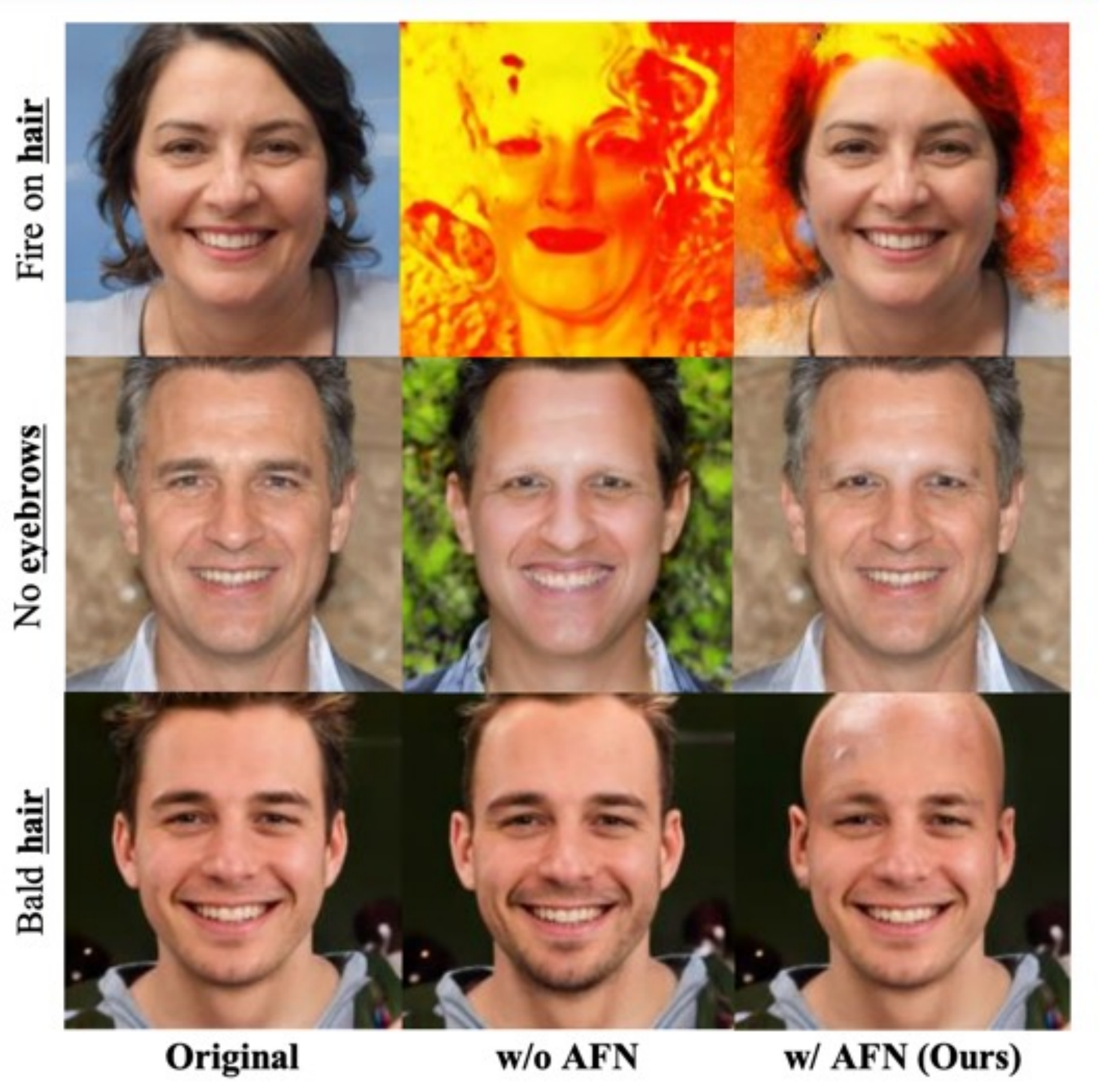}
%\vspace{-1.5em}
\caption{Results of out-of-distribution cases.}
\label{fig: ood}
%\vspace{-1.0em}
\end{figure}

\noindent\textbf{3D Distillation of CLIP's Knowledge} As we lack ground-truth 3D semantic masks to supervise the 3D attention field, we render them into 2D and supervise them using generated 2D pseudo-masks. 
This is necessary since training AFN directly with $\mathcal{L}_{CLIP^{+}}$ often leads to the generation of irrelevant and poorly localized attention fields, resulting in severe artifacts~(Please see supplement). This is not surprising, given that the target text prompt $t_{edit}$ does not always align with the region of interest, and the CLIP signal tends to have a global effect on the radiance fields. Furthermore, we presume that additional complexity to the model without adequate regularization easily leads to degenerate solutions. 

Therefore we first introduce a separate masking prompt $t_{mask}$ solely describing our region of interest. Next, we utilize $t_{mask}$ together with the CLIP model to generate a 2D relevance mask of the source or target image that serves as a pseudo-label for AFN training. Inspired by transformer visualization methods~\cite{DBLP:conf/cvpr/CheferGW21, bach2015pixel}, we aggregate relevance maps across the transformer heads and layers to build a relevance mask $\mathbf{M} \in \mathbb{R} ^ {H_{V} \times W_{V}}$ with respect to the model output $y$, given $t_{mask}$ (More details in the supplementary).
%Refer to supplement for details).

Unlike~\cite{DBLP:conf/cvpr/CheferGW21}, where the transformer output $y_t$ is the logit value of class $t$, our output $y$ is the cosine similarity between the embedding vector $E_{t}(t_{mask})$ and $E_{I}(I)$.
Generally, it is sufficient to use the source image $I_s$ for the mask generation, but for large deformations, we estimate two masks from both the source and intermediate images $I_s$ and $I_t$ and perform a $\max$ operation.

%The key idea of this approach is to generate robust relevance mask given $t_{mask}$ by integrating the relevance map of many layers of the transformer encoder, while taking into account the characteristics of transformer block such as skip connections and non-linearites other than ReLU that could output negative values. 

%E_t, E_I 정의 해줘야함
% I 는 rendered img of source radiance 일 수도 있고, large deformation 이 필요한 경우에는 raw_edit radiance의 rendered img를 기준으로 enlarged mask를 예측하도록 학습할 수 있음

% \hat{\mathbf{M}} = {\{ \pi_{vol}({t_{jk} \}}^{N_s}_{j=1}) \}}^{H_{V} \times W_{V}}_{k=1}

The estimated relevance mask $\mathbf{M}$ is used as a pseudo-label to guide the training of AFN by minimizing the MSE loss $\mathcal{L}_{mask}$ between the volume-rendered attention field $\hat{\mathbf{M}}_t$ and $\mathbf{M}$. %($\mathcal{L}_{mask} = {\lVert \mathbf{M} -  \hat{\mathbf{M}} \rVert_{2}^{2}}$).
% \end{aligned}
% \end{equation}
% \begin{equation}
% \begin{aligned}
% \mathcal{L}_{mask} = {\lVert \mathbf{M} -  \hat{\mathbf{M}} \rVert_{2}^{2}},
% \end{aligned}
% \end{equation}
%where $N_s$ is the number of sample points per ray. 
While the generated pseudo-label $\mathbf{M}$ is low-dimensional and often inaccurate, the semantic information contained in the input features to AFN allows for training even with noisy guidance~(Figure~\ref{fig_main} (d)). 
\\
%We experimentally show that it leads to convergence of 3D attention fields only using 2D noisy pseudo-label. \textcolor{red}{Figure}

% Q. loss 에 expectation 취해줘야하나??
% Loss 

\begin{figure}[t!]
\centering
\includegraphics[width=1.0\linewidth]{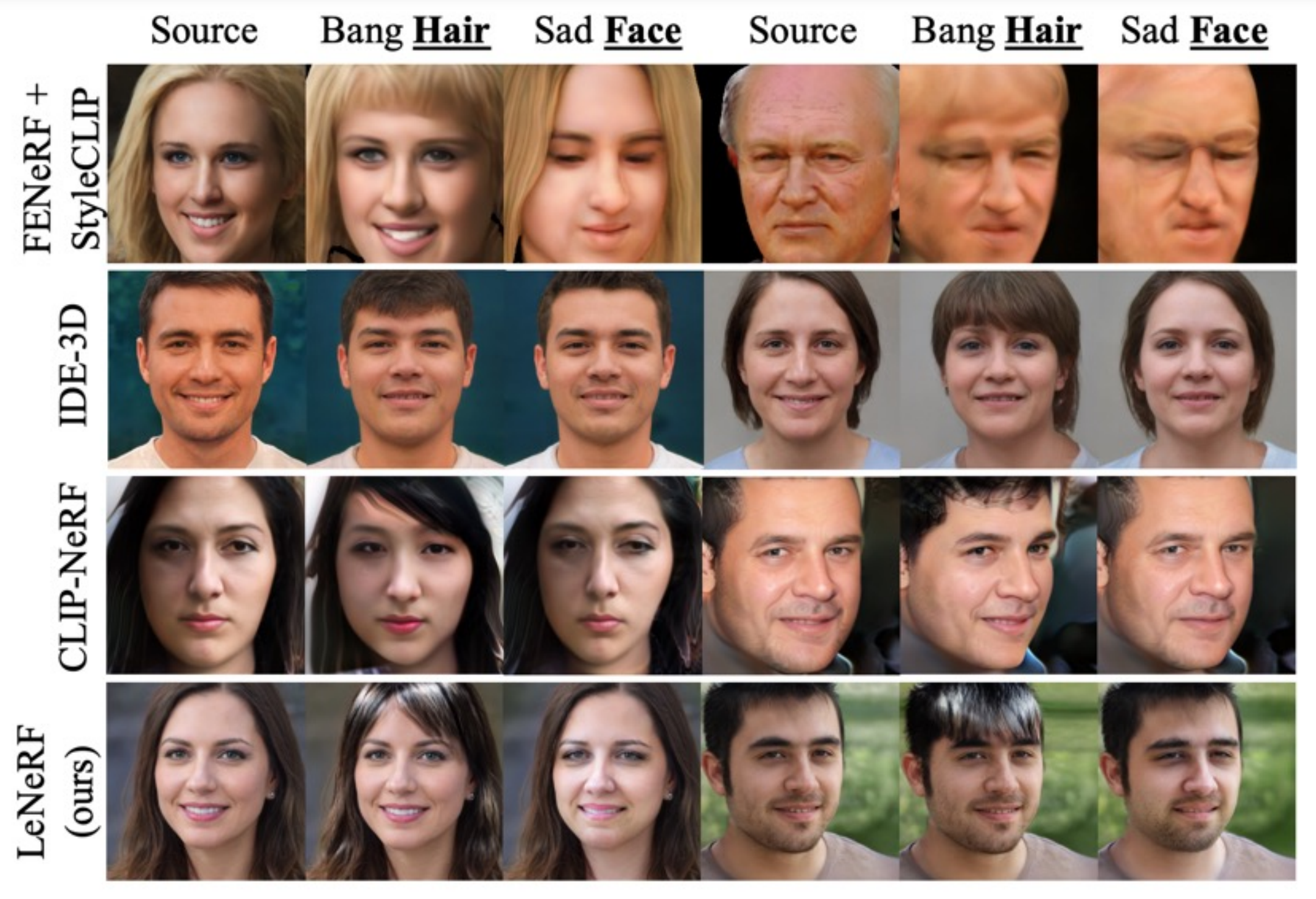}
%\vspace{-1.5em}
\caption{Results of \ours and various baselines.}
\label{fig: comparison}
%\vspace{-1.0em}
\end{figure}

\noindent\textbf{Mask Regularization} We utilize total variation regularization~\cite{DBLP:conf/icip/RudinO94} (Eq. \ref{eq:tv}) to regularize the 3D attention field for spatial smoothness, where $\Delta_{x}^{2}m$ denotes the squared difference between $m_{i+1,j,k}$ and $m_{i,j,k}$, and analogously for $\Delta_{y}^{2}m$ and $\Delta_{z}^{2}m$:

\begin{equation} \label{eq:tv}
\begin{aligned}
\mathcal{L}_{tv} = \frac{1}{N} \sum_{m \in \mathcal{A}}{\sqrt{\Delta_{x}^{2}m + \Delta_{y}^{2}m + \Delta_{z}^{2}m }}.
\end{aligned}
\end{equation}
Also, the sparsity regularization (Eq. \ref{eq:sparse}) is used in order to focus on the region of interest:
\begin{equation} \label{eq:sparse}
\begin{aligned}
\mathcal{L}_{sparsity} = -\sum_{\mathbf{x}_i \in K_{top}} {\log{m_i}} - \sum_{\mathbf{x}_i \in K_{bottom}}{log{(1 - m_{i})}},
\end{aligned}
\end{equation}
where $K_{top}$ refers to the set of coordinates with the top-$k$ $m$ values, and $K_{bottom}$ refers to the bottom-$k$. In summary, the objective function for AFN $A$ is
\begin{equation}
\begin{aligned}
\mathcal{L}_{AFN} = \lambda_{mask}\mathcal{L}_{mask} + \lambda_{tv}\mathcal{L}_{tv} \\
+ \lambda_{sparsity}\mathcal{L}_{sparsity} + \lambda_{CLIP^+}\mathcal{L}_{CLIP^+}.
\end{aligned}
\end{equation}

\noindent Please refer to the supplement for additional descriptions, hyperparameters, and implementation details.
%our final loss function for $M_{\theta_{m}}$ and $F_{\theta_{f}}$ are:

%\input{05_Experiments}

\section{Experiments}

%\subsection{Setup}
\noindent\textbf{Datasets}
LeNeRF utilizes pretrained EG3D~\cite{EG3D} generator trained on FFHQ~\cite{DBLP:conf/cvpr/KarrasLA19}, AFHQv2 CATS~\cite{DBLP:conf/cvpr/ChoiUYH20}, and ShapeNet Cars~\cite{DBLP:conf/cvpr/ChenZ19, DBLP:conf/nips/LiuGLCT20}.
%and shows that the method is generally applicable many domains. 
However, please note that the datasets are not used when training \ours.

% \subsection{Comparisons.}
\noindent\textbf{Baselines} We compare \ours against three state-of-the-art NeRF editing methods: CLIP-NeRF~\cite{DBLP:conf/cvpr/WangCH0022}\footnote{Note that the full code of CLIP-NeRF has not been released at the time of writing. Therefore, the results are our reproduced CLIP-NeRF.}, FENeRF~\cite{Sun_2022_CVPR}, and IDE-3D~\cite{DBLP:journals/corr/abs-2205-15517}. 
%We re-implement CLIP-NeRF since the full code is not available. 
While CLIP-NeRF is a text-guided editing model, FENeRF and IDE-3D are not directly comparable to our method since semantic masks are used as a control handle. Therefore we create two new baselines based on the two models (FENeRF+StyleCLIP and IDE-3D+StyleCLIP) which can edit radiance fields with text guidance. Specifically, we invert the images generated by the two methods to StyleGAN latent codes using e4e~\cite{tov2021designing} encoder. StyleCLIP~\cite{Patashnik_2021_ICCV} is used to edit the inverted images given a target text prompt. Face semantic maps are then extracted from the edited images using a pretrained face parsing network of SofGAN~\cite{DBLP:journals/tog/ChenLXCSY22}, which is used for the input guidance. Note that only FFHQ results can be reported for FENeRF and IDE-3D since there are no available semantic labels and pretrained parsing networks for other datasets.

\begin{figure}[t!]
\centering
\includegraphics[width=1.0\linewidth]{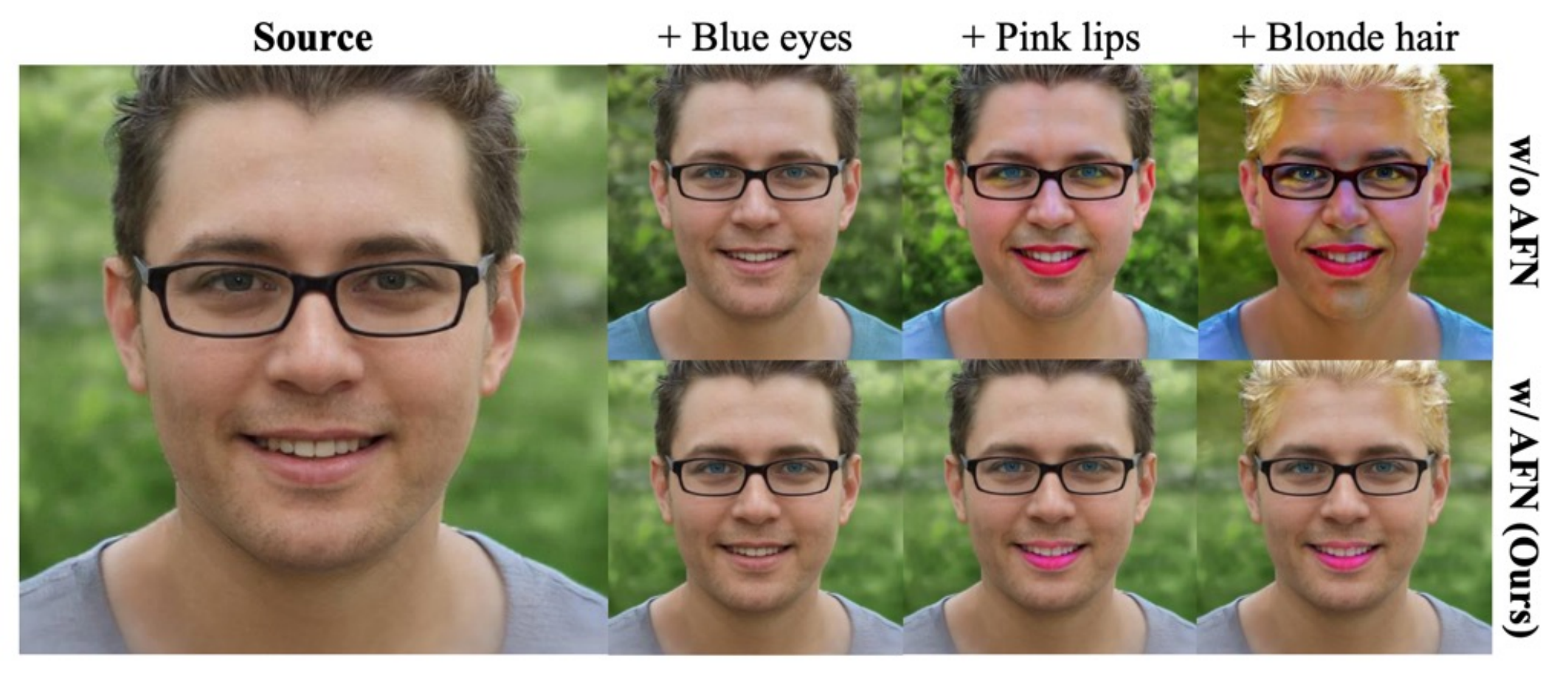}
%\vspace{-0.1in}
\caption{Result of sequential editing.}
%Source and editing results with pixel difference maps. LENeRF performs localized editing while LENeRF without AFN suffers from global and unwanted changes.}
\label{fig: sequential}
\end{figure}

%\subsection{Results}
\noindent\textbf{Qualitative evaluations} Fig.~\ref{fig:curated} provides curated results of our model, showing the quality, diversity, and view consistency of \ours. Fig.~\ref{fig: comparison} presents qualitative comparisons against other three methods. 
FENeRF+StyleCLIP and CLIP-NeRF fail to generate high-quality renderings, and the quality of images degrades even further after editing. 
IDE-3D+StyleCLIP synthesizes high-fidelity images before editing, but it is difficult to edit texture or expression using semantic mask guidance. Also, semantic mask-based methods require either inversion steps which cannot be done in real-time or an encoding step which degrades identity preservation. \ours, however, can perform localized editing in real-time and preserve the quality and the identity of original content.

Fig.~\ref{fig: partial_editing} shows partial editing results where we use the same attribute prompt $t_{edit}$ while varying the mask prompt $t_{mask}$. This enables highly localized editing without unwanted changes or color-bleeding artifacts. 
Fig.~\ref{fig: sequential} demonstrates sequential editing results where we perform different manipulations sequentially. Multiple changes to the source latent code destroy the initial content gradually when trained without \afn (bottom) whereas, the identity is preserved in \ours (top) by manipulating only the local parts of the radiance field. 
Fig.~\ref{fig: ood} shows that \ours is robust to out-of-distribution editing (e.g., fire on \underline{hair}). \ours has the representation capability to generate samples that are outside of the latent space by fusing the two radiance fields, thereby reflecting the target prompts while preserving the identity.
%Real Portrait Editing inversion - supplement?

\begin{table}[t!]
\centering
% \fontsize{7.0}{10}\selectfont

\caption{PSNR, R-precision, and FID performance of various baselines and \ours on FFHQ and Cats.}
\vspace{-0.1in}
\label{table_all}
\begin{adjustbox}{width=1\linewidth}
\vspace{-1.0em}
\scriptsize
\setlength{\tabcolsep}{3pt}
\begin{tabular}[\linewidth]{c cccc cc}
\toprule
 & \multicolumn{4}{c}{FFHQ} & \multicolumn{2}{c}{Cats}  \\
 \cmidrule(lr){2-5} \cmidrule(lr){6-7}
      
 & PSNR $\uparrow$  & R-pre $\uparrow$ & FID $\downarrow$ & $\Delta$FID $\downarrow$ &  FID $\downarrow$& $\Delta$FID $\downarrow$   \\
\midrule
CLIP-NeRF &
7.08 & 0.21 & 41.5 & +19.8  & 18.6 & +12.7  \\
FeNeRF + SC &
5.33 & 0.19 & 37.4 & +31.2 & - & -   \\
IDE3D + SC &
12.44 & 0.61 & 4.56 & +3.29 & - & -   \\
LeNeRF w/o AFN &
10.81 & 0.66 & $\mathbf{4.37}$ & +4.90 & $\mathbf{2.71}$ & +4.17  \\
\textbf{LeNeRF (Ours)} &
$\mathbf{20.86}$ & $\mathbf{0.78}$ & $\mathbf{4.37}$ & +$\mathbf{2.21}$ & $\mathbf{2.71}$ & +$\mathbf{1.94}$  \\
\bottomrule
\vspace{-2.0em}
\end{tabular}
\end{adjustbox}
\end{table}

\begin{figure}[t!]
\centering
\includegraphics[width=0.9\linewidth]{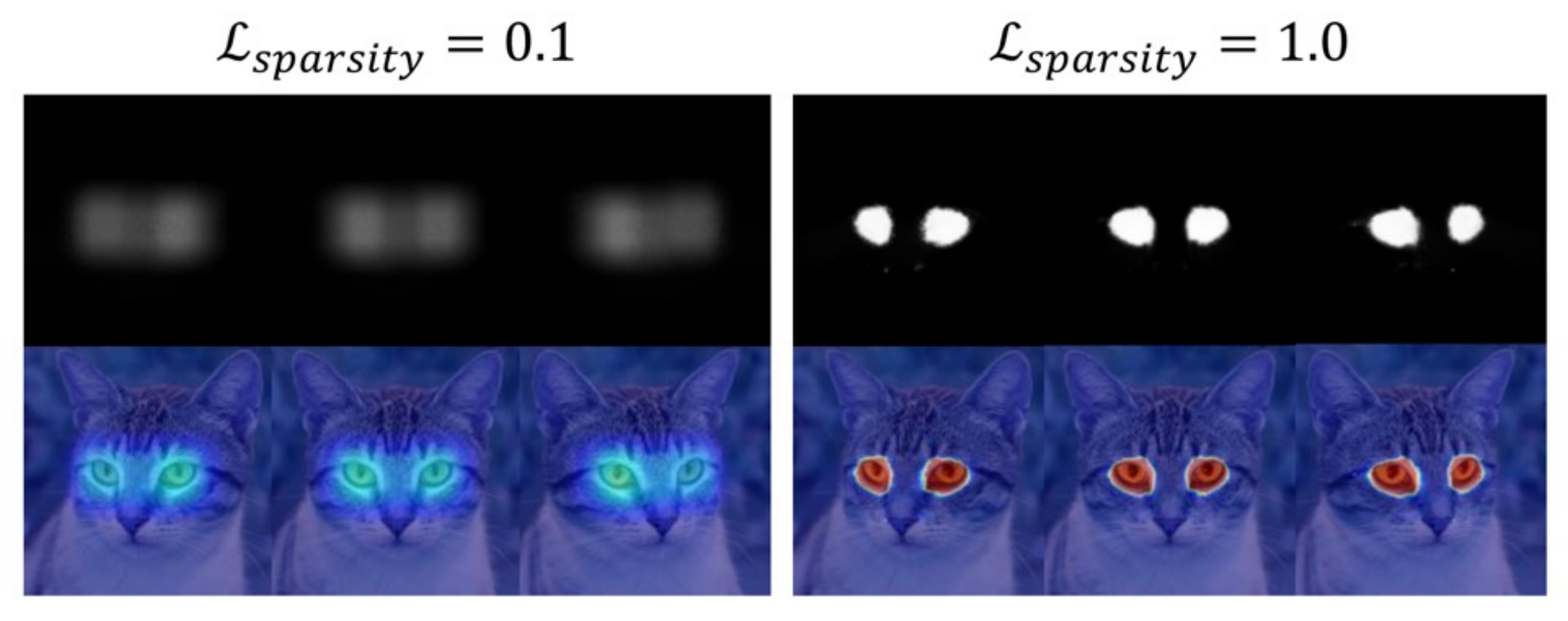}
%\vspace{-0.1in}
\caption{Results of the effect of $\lambda_{sparsity}$ on the mask.}
%(left) and high $\lambda_{sparsity}$ (right). Results with high $\lambda_{sparsity}$ show a clear mask.}
\label{fig: sparsity}
%\vspace{-1.0em}
\end{figure}

% 우리꺼 cat inference중 => 되는대로 cat FID 측정 
% fenerf:  lenerf/evaluation 에 올려놓음. FID 돌리기
% SHAPENET cars CLIP-NeRF FID 측정: pi-GAN/shapenet_results 폴더에 생성중
% lambda x : -10. * torch.log(x) / torch.log(torch.Tensor([10.])) psnr 계산

\noindent\textbf{Quantitative evaluations} Table~\ref{table_all} provides quantitative measures of \ours and various baselines.
We quantify the unintended changes by computing the PSNR between the source and the manipulated images outside of the region of interest specified by $t_{mask}$. The ground-truth mask of the region is detected by a pretrained face parsing network\cite{DBLP:journals/tog/ChenLXCSY22}. We assess the quality of the source images and manipulated images using Fr\'{e}chet Inception Distance (FID)~\cite{DBLP:conf/nips/HeuselRUNH17}, and compare the differences to evaluate post-editing degradation. 
Also for FFHQ, we report R-precision using a pretrained attribute classifier~\cite{DBLP:conf/iccv/LiuLWT15} to measure how well the text prompt is reflected in the manipulated results. 
We generate 30k samples with randomly sampled text prompts to calculate FID, and sample 1K images per text prompt to calculate PSNR and R-precision.
%\ours demonstrates the best quality after the manipulation, and also the smallest FID difference before and after editing indicates the robustness of our method. Also, our method shows the best performance in achieving the desired manipulations and minimizing unwanted changes. 

Among the models, \ours shows the best FID score, demonstrating that it produces high-quality results after the manipulation. 
Also, \ours have the smallest FID difference between the images before and after the edit, indicating the robustness of the model. 
Moreover, \ours shows the highest PSNR and R-precision; it has the best performance in achieving the desired manipulations while minimizing unintended changes.
Please refer to the supplement for more details.

\noindent\textbf{User study} We ask users to evaluate \ours along with various baselines in the range from 1 to 10 regarding 1) fidelity, 2) locality, 3) identity preservation, and 4) how well the text prompt is reflected in the results. 
%Users scored each criterion in the range from 0 to 10, and LENeRF outperforms all baselines by a large margin.
\ours outperforms all baselines by a large margin on each criterion, and the scores are in the supplement.

\begin{figure}[t!]
\centering
\includegraphics[width=1.0\linewidth]{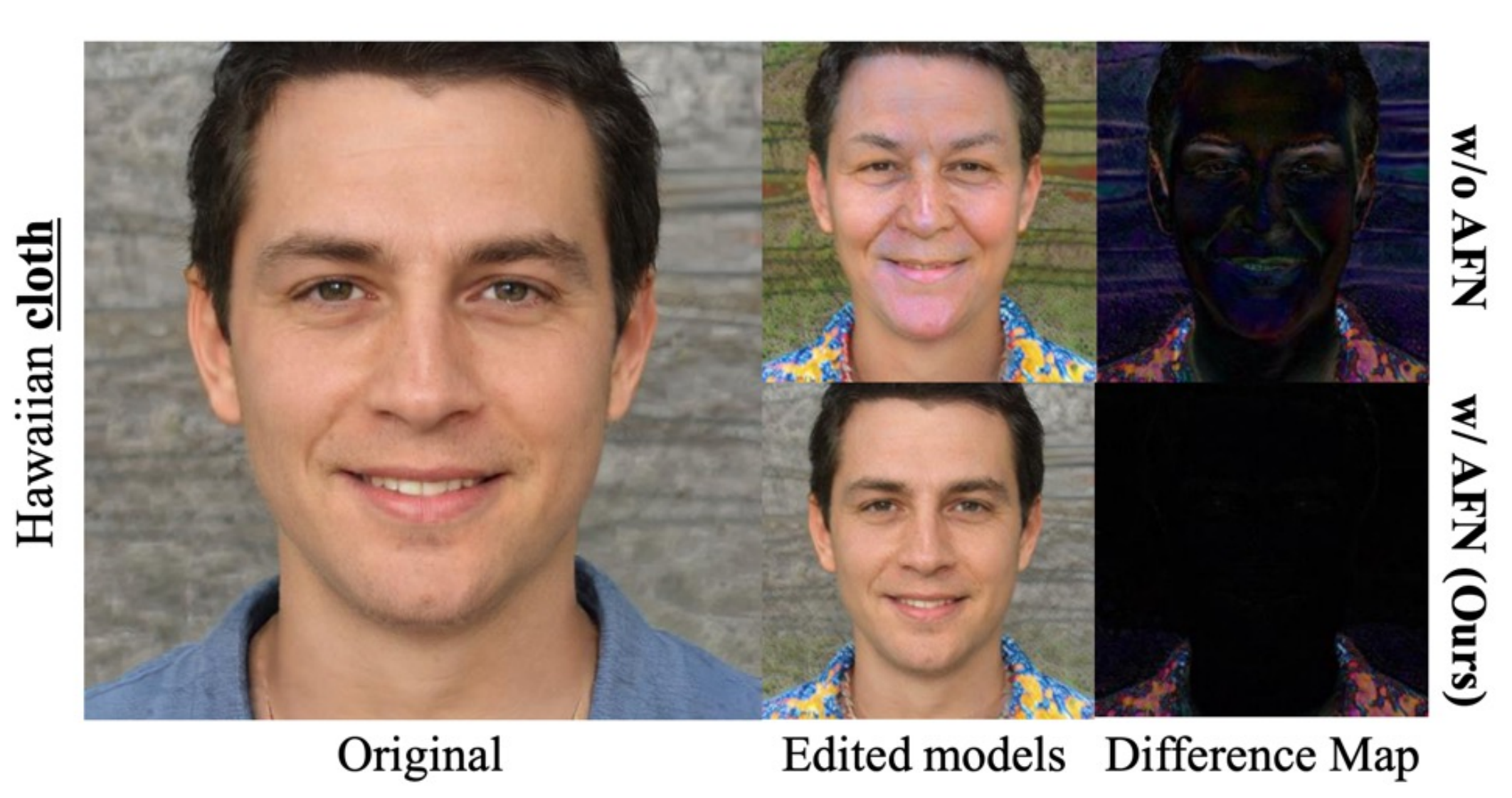}
%\vspace{-0.1in}
\caption{Results on comparison of pixel-wise difference maps.}
%Source and editing results with pixel difference maps. LENeRF performs localized editing while LENeRF without AFN suffers from global and unwanted changes.}
\label{fig: diff_map}
\end{figure}

\begin{figure}[t!]
\centering
\includegraphics[width=1.0\linewidth]{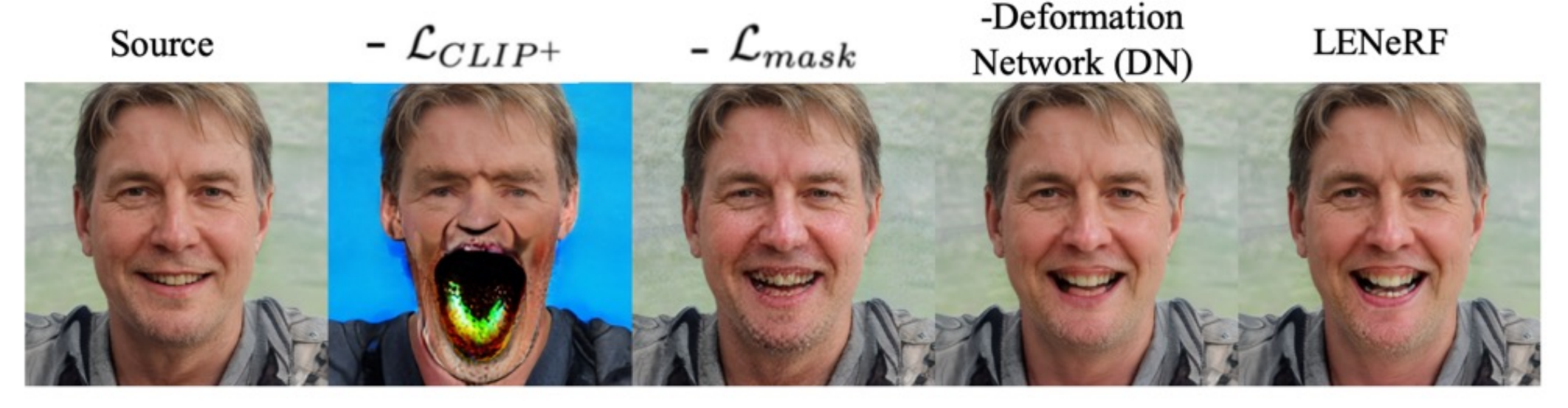}
%\vspace{-0.1in}
\caption{Importance of our objective functions $\mathcal{L}_{CLIP^{+}}$, $\mathcal{L}_{mask}$, and Deformation Network~(\dn).}
\label{fig: loss}
\end{figure}

\noindent\textbf{Ablation study}
Fig.~\ref{fig: sparsity} shows the effect of controlling $\lambda_{sparsity}$. A larger weight on $\mathcal{L}_{sparsity}$ produces a sharper 3D mask whereas a smaller weight leads to a smoother mask.
Fig.~\ref{fig: diff_map} visualizes the pixel-wise difference map to show the effect of \afn and the radiance fusion technique on the locality. Generating a 3D mask helps on minimize the unintended changes outside of the region of interest and reduces color-bleeding artifacts that are incorporated with the target prompt $t_{edit}$.
Fig.~\ref{fig: loss} shows the importance of the \dn and our objective functions $\mathcal{L}_{CLIP^{+}}$ and $\mathcal{L}_{mask}$. Minimizing the naive CLIP loss results in degenerate solutions. Training \afn without $\mathcal{L}_{mask}$ fails to estimate a 3D mask, resulting in severe artifacts.
Finally, \ours without \dn leads to the interpolation of unrelated features and cannot handle geometric manipulations.

\section{Conclusion}

% \noindent\textbf{Ethical consideration} We recognize that our method is biased towards the pretrained generator and the CLIP model. We also consider the potential misuse of our model such as spreading misinformation intentionally and are willing to provide safety features before the code release.
% \\

\noindent\textbf{Conclusion} While texts provide rich and descriptive information for editing, they often lack locality. \ours focuses on this aspect and proposes a point-wise feature manipulation technique via 3D mask generation and radiance fusion. We show that 3D-aware knowledge of NeRF and rich multi-modal information from CLIP can be combined to create a robust 3D mask without any additional datasets.
\ours generates high-fidelity and find-grained 3D editing results. 
%Due to limited pages, we add our limitations on the supplementary page.

%\noindent\textbf{Limitations and future work.} The main limitation of our method is dependent on the capability of the pretrained 3D generator and the CLIP model. Due to limited pages, we add our limitations on the supplementary page.

% \noindent\textbf{Limitations and future work.} Our editing method is dependent on the capability of the pretrained 3D generator and the CLIP model. The method suffers from synthesizing samples outside of the latent space of the generator, so the diversity of the editing is limited. We often find artifacts for edits with large deformations, since the CLIP loss easily leads to degenerate solutions. One way is to use a stronger 3D generator, and the other way is to work on replacing the CLIP model with models that have stronger priors such as diffusion-based text to image generation models~\cite{rombach2021highresolution,DBLP:journals/corr/abs-2205-11487,DBLP:journals/corr/abs-2204-06125}, which have shown promising results in multi-modal understanding and generation and might help enhance the diversity and quality of 3D editing. Also, text guidance is very descriptive but is unable to guide the degree of editing (e.g., how much to open the mouth). External control handles such as 3D morphable models (3DMM) can be utilized for such control.
% \\

% Acknowledgement 관련 명령어 없는듯? 일단 따로 작성해놨음.
{
\small
\noindent\textbf{Acknowledgments.}
This work was supported by the Institute of Information \& communications Technology Planning \& Evaluation (IITP) grant funded by the Korean government (MSIT) (No. 2019-0-00075, Artificial Intelligence Graduate School Program (KAIST) and No. 2021-0-01778, Development of human image synthesis and discrimination technology below the perceptual threshold).
}

%%%%%%%%% REFERENCES
{\small
\bibliographystyle{ieee_fullname}
\bibliography{main}

\begin{thebibliography}{10}\itemsep=-1pt

\bibitem{DBLP:conf/siggraph/AbdalZ0MW22}
Rameen Abdal, Peihao Zhu, John Femiani, Niloy~J. Mitra, and Peter Wonka.
\newblock Clip2stylegan: Unsupervised extraction of stylegan edit directions.
\newblock In Munkhtsetseg Nandigjav, Niloy~J. Mitra, and Aaron Hertzmann,
  editors, {\em {SIGGRAPH} '22: Special Interest Group on Computer Graphics and
  Interactive Techniques Conference, Vancouver, BC, Canada, August 7 - 11,
  2022}, pages 48:1--48:9. {ACM}, 2022.

\bibitem{bach2015pixel}
Sebastian Bach, Alexander Binder, Gr{\'e}goire Montavon, Frederick Klauschen,
  Klaus-Robert M{\"u}ller, and Wojciech Samek.
\newblock On pixel-wise explanations for non-linear classifier decisions by
  layer-wise relevance propagation.
\newblock {\em PloS one}, 10(7):e0130140, 2015.

\bibitem{DBLP:journals/corr/abs-2202-06079}
Zehranaz Canfes, M.~Furkan Atasoy, Alara Dirik, and Pinar Yanardag.
\newblock Text and image guided 3d avatar generation and manipulation.
\newblock {\em CoRR}, abs/2202.06079, 2022.

\bibitem{EG3D}
Eric~R. Chan, Connor~Z. Lin, Matthew~A. Chan, Koki Nagano, Boxiao Pan,
  Shalini~De Mello, Orazio Gallo, Leonidas~J. Guibas, Jonathan Tremblay, Sameh
  Khamis, Tero Karras, and Gordon Wetzstein.
\newblock Efficient geometry-aware 3d generative adversarial networks.
\newblock In {\em {IEEE/CVF} Conference on Computer Vision and Pattern
  Recognition, {CVPR} 2022, New Orleans, LA, USA, June 18-24, 2022}, pages
  16102--16112. {IEEE}, 2022.

\bibitem{DBLP:conf/cvpr/ChanMK0W21}
Eric~R. Chan, Marco Monteiro, Petr Kellnhofer, Jiajun Wu, and Gordon Wetzstein.
\newblock Pi-gan: Periodic implicit generative adversarial networks for
  3d-aware image synthesis.
\newblock In {\em {IEEE} Conference on Computer Vision and Pattern Recognition,
  {CVPR} 2021, virtual, June 19-25, 2021}, pages 5799--5809. Computer Vision
  Foundation / {IEEE}, 2021.

\bibitem{DBLP:conf/cvpr/CheferGW21}
Hila Chefer, Shir Gur, and Lior Wolf.
\newblock Transformer interpretability beyond attention visualization.
\newblock In {\em {IEEE} Conference on Computer Vision and Pattern Recognition,
  {CVPR} 2021, virtual, June 19-25, 2021}, pages 782--791. Computer Vision
  Foundation / {IEEE}, 2021.

\bibitem{DBLP:journals/tog/ChenLXCSY22}
Anpei Chen, Ruiyang Liu, Ling Xie, Zhang Chen, Hao Su, and Jingyi Yu.
\newblock Sofgan: {A} portrait image generator with dynamic styling.
\newblock {\em {ACM} Trans. Graph.}, 41(1):1:1--1:26, 2022.

\bibitem{DBLP:conf/cvpr/ChenZ19}
Zhiqin Chen and Hao Zhang.
\newblock Learning implicit fields for generative shape modeling.
\newblock In {\em {IEEE} Conference on Computer Vision and Pattern Recognition,
  {CVPR} 2019, Long Beach, CA, USA, June 16-20, 2019}, pages 5939--5948.
  Computer Vision Foundation / {IEEE}, 2019.

\bibitem{DBLP:conf/cvpr/ChoiUYH20}
Yunjey Choi, Youngjung Uh, Jaejun Yoo, and Jung{-}Woo Ha.
\newblock Stargan v2: Diverse image synthesis for multiple domains.
\newblock In {\em 2020 {IEEE/CVF} Conference on Computer Vision and Pattern
  Recognition, {CVPR} 2020, Seattle, WA, USA, June 13-19, 2020}, pages
  8185--8194. Computer Vision Foundation / {IEEE}, 2020.

\bibitem{DBLP:journals/pami/DengGYXKZ22}
Jiankang Deng, Jia Guo, Jing Yang, Niannan Xue, Irene Kotsia, and Stefanos
  Zafeiriou.
\newblock Arcface: Additive angular margin loss for deep face recognition.
\newblock {\em {IEEE} Trans. Pattern Anal. Mach. Intell.}, 44(10):5962--5979,
  2022.

\bibitem{DBLP:conf/cvpr/DengYX022}
Yu Deng, Jiaolong Yang, Jianfeng Xiang, and Xin Tong.
\newblock {GRAM:} generative radiance manifolds for 3d-aware image generation.
\newblock In {\em {IEEE/CVF} Conference on Computer Vision and Pattern
  Recognition, {CVPR} 2022, New Orleans, LA, USA, June 18-24, 2022}, pages
  10663--10673. {IEEE}, 2022.

\bibitem{DBLP:journals/tog/GalPMBCC22}
Rinon Gal, Or Patashnik, Haggai Maron, Amit~H. Bermano, Gal Chechik, and Daniel
  Cohen{-}Or.
\newblock Stylegan-nada: Clip-guided domain adaptation of image generators.
\newblock {\em {ACM} Trans. Graph.}, 41(4):141:1--141:13, 2022.

\bibitem{DBLP:conf/cvpr/GatysEB16}
Leon~A. Gatys, Alexander~S. Ecker, and Matthias Bethge.
\newblock Image style transfer using convolutional neural networks.
\newblock In {\em 2016 {IEEE} Conference on Computer Vision and Pattern
  Recognition, {CVPR} 2016, Las Vegas, NV, USA, June 27-30, 2016}, pages
  2414--2423. {IEEE} Computer Society, 2016.

\bibitem{DBLP:conf/cvpr/GenovaCSSF20}
Kyle Genova, Forrester Cole, Avneesh Sud, Aaron Sarna, and Thomas~A.
  Funkhouser.
\newblock Local deep implicit functions for 3d shape.
\newblock In {\em 2020 {IEEE/CVF} Conference on Computer Vision and Pattern
  Recognition, {CVPR} 2020, Seattle, WA, USA, June 13-19, 2020}, pages
  4856--4865. Computer Vision Foundation / {IEEE}, 2020.

\bibitem{GoodfellowPMXWOCB14}
Ian~J. Goodfellow, Jean Pouget{-}Abadie, Mehdi Mirza, Bing Xu, David
  Warde{-}Farley, Sherjil Ozair, Aaron~C. Courville, and Yoshua Bengio.
\newblock Generative adversarial nets.
\newblock In {\em Proc. the Advances in Neural Information Processing Systems
  (NeurIPS)}, 2014.

\bibitem{DBLP:conf/iclr/GuL0T22}
Jiatao Gu, Lingjie Liu, Peng Wang, and Christian Theobalt.
\newblock Stylenerf: {A} style-based 3d aware generator for high-resolution
  image synthesis.
\newblock In {\em The Tenth International Conference on Learning
  Representations, {ICLR} 2022, Virtual Event, April 25-29, 2022}.
  OpenReview.net, 2022.

\bibitem{DBLP:conf/iccv/GuillardRYF21}
Beno{\^{\i}}t Guillard, Edoardo Remelli, Pierre Yvernay, and Pascal Fua.
\newblock Sketch2mesh: Reconstructing and editing 3d shapes from sketches.
\newblock In {\em 2021 {IEEE/CVF} International Conference on Computer Vision,
  {ICCV} 2021, Montreal, QC, Canada, October 10-17, 2021}, pages 13003--13012.
  {IEEE}, 2021.

\bibitem{DBLP:conf/nips/HeuselRUNH17}
Martin Heusel, Hubert Ramsauer, Thomas Unterthiner, Bernhard Nessler, and Sepp
  Hochreiter.
\newblock Gans trained by a two time-scale update rule converge to a local nash
  equilibrium.
\newblock In Isabelle Guyon, Ulrike von Luxburg, Samy Bengio, Hanna~M. Wallach,
  Rob Fergus, S.~V.~N. Vishwanathan, and Roman Garnett, editors, {\em Advances
  in Neural Information Processing Systems 30: Annual Conference on Neural
  Information Processing Systems 2017, December 4-9, 2017, Long Beach, CA,
  {USA}}, pages 6626--6637, 2017.

\bibitem{DBLP:journals/tog/HongZPCYL22}
Fangzhou Hong, Mingyuan Zhang, Liang Pan, Zhongang Cai, Lei Yang, and Ziwei
  Liu.
\newblock Avatarclip: zero-shot text-driven generation and animation of 3d
  avatars.
\newblock {\em {ACM} Trans. Graph.}, 41(4):161:1--161:19, 2022.

\bibitem{DBLP:conf/cvpr/JainMBAP22}
Ajay Jain, Ben Mildenhall, Jonathan~T. Barron, Pieter Abbeel, and Ben Poole.
\newblock Zero-shot text-guided object generation with dream fields.
\newblock In {\em {IEEE/CVF} Conference on Computer Vision and Pattern
  Recognition, {CVPR} 2022, New Orleans, LA, USA, June 18-24, 2022}, pages
  857--866. {IEEE}, 2022.

\bibitem{DBLP:conf/cvpr/JiangSMHNF20}
Chiyu~"Max" Jiang, Avneesh Sud, Ameesh Makadia, Jingwei Huang, Matthias
  Nie{\ss}ner, and Thomas~A. Funkhouser.
\newblock Local implicit grid representations for 3d scenes.
\newblock In {\em 2020 {IEEE/CVF} Conference on Computer Vision and Pattern
  Recognition, {CVPR} 2020, Seattle, WA, USA, June 13-19, 2020}, pages
  6000--6009. Computer Vision Foundation / {IEEE}, 2020.

\bibitem{DBLP:conf/cvpr/KaniaYKTT22}
Kacper Kania, Kwang~Moo Yi, Marek Kowalski, Tomasz Trzcinski, and Andrea
  Tagliasacchi.
\newblock Conerf: Controllable neural radiance fields.
\newblock In {\em {IEEE/CVF} Conference on Computer Vision and Pattern
  Recognition, {CVPR} 2022, New Orleans, LA, USA, June 18-24, 2022}, pages
  18602--18611. {IEEE}, 2022.

\bibitem{DBLP:conf/cvpr/KarrasLA19}
Tero Karras, Samuli Laine, and Timo Aila.
\newblock A style-based generator architecture for generative adversarial
  networks.
\newblock In {\em {IEEE} Conference on Computer Vision and Pattern Recognition,
  {CVPR} 2019, Long Beach, CA, USA, June 16-20, 2019}, pages 4401--4410.
  Computer Vision Foundation / {IEEE}, 2019.

\bibitem{Karras2019stylegan2}
Tero Karras, Samuli Laine, Miika Aittala, Janne Hellsten, Jaakko Lehtinen, and
  Timo Aila.
\newblock Analyzing and improving the image quality of {StyleGAN}.
\newblock In {\em Proc. CVPR}, 2020.

\bibitem{KimKC21}
Daejin Kim, Mohammad~Azam Khan, and Jaegul Choo.
\newblock Not just compete, but collaborate: Local image-to-image translation
  via cooperative mask prediction.
\newblock In {\em Proc. of the IEEE conference on computer vision and pattern
  recognition (CVPR)}, 2021.

\bibitem{DBLP:journals/corr/abs-2205-15585}
Sosuke Kobayashi, Eiichi Matsumoto, and Vincent Sitzmann.
\newblock Decomposing nerf for editing via feature field distillation.
\newblock {\em CoRR}, abs/2205.15585, 2022.

\bibitem{DBLP:conf/icml/KosiorekSZMSMR21}
Adam~R. Kosiorek, Heiko Strathmann, Daniel Zoran, Pol Moreno, Rosalia
  Schneider, Sona Mokr{\'{a}}, and Danilo~Jimenez Rezende.
\newblock Nerf-vae: {A} geometry aware 3d scene generative model.
\newblock In Marina Meila and Tong Zhang, editors, {\em Proceedings of the 38th
  International Conference on Machine Learning, {ICML} 2021, 18-24 July 2021,
  Virtual Event}, volume 139 of {\em Proceedings of Machine Learning Research},
  pages 5742--5752. {PMLR}, 2021.

\bibitem{DBLP:conf/nips/LiuGLCT20}
Lingjie Liu, Jiatao Gu, Kyaw~Zaw Lin, Tat{-}Seng Chua, and Christian Theobalt.
\newblock Neural sparse voxel fields.
\newblock In Hugo Larochelle, Marc'Aurelio Ranzato, Raia Hadsell,
  Maria{-}Florina Balcan, and Hsuan{-}Tien Lin, editors, {\em Advances in
  Neural Information Processing Systems 33: Annual Conference on Neural
  Information Processing Systems 2020, NeurIPS 2020, December 6-12, 2020,
  virtual}, 2020.

\bibitem{DBLP:conf/iccv/LiuZZ0ZR21}
Steven Liu, Xiuming Zhang, Zhoutong Zhang, Richard Zhang, Jun{-}Yan Zhu, and
  Bryan Russell.
\newblock Editing conditional radiance fields.
\newblock In {\em 2021 {IEEE/CVF} International Conference on Computer Vision,
  {ICCV} 2021, Montreal, QC, Canada, October 10-17, 2021}, pages 5753--5763.
  {IEEE}, 2021.

\bibitem{DBLP:conf/iccv/LiuLWT15}
Ziwei Liu, Ping Luo, Xiaogang Wang, and Xiaoou Tang.
\newblock Deep learning face attributes in the wild.
\newblock In {\em 2015 {IEEE} International Conference on Computer Vision,
  {ICCV} 2015, Santiago, Chile, December 7-13, 2015}, pages 3730--3738. {IEEE}
  Computer Society, 2015.

\bibitem{DBLP:conf/iccv/MehrJTCG19}
{\'{E}}loi Mehr, Ariane Jourdan, Nicolas Thome, Matthieu Cord, and Vincent
  Guitteny.
\newblock Disconet: Shapes learning on disconnected manifolds for 3d editing.
\newblock In {\em 2019 {IEEE/CVF} International Conference on Computer Vision,
  {ICCV} 2019, Seoul, Korea (South), October 27 - November 2, 2019}, pages
  3473--3482. {IEEE}, 2019.

\bibitem{DBLP:conf/cvpr/MichelBLBH22}
Oscar Michel, Roi Bar{-}On, Richard Liu, Sagie Benaim, and Rana Hanocka.
\newblock Text2mesh: Text-driven neural stylization for meshes.
\newblock In {\em {IEEE/CVF} Conference on Computer Vision and Pattern
  Recognition, {CVPR} 2022, New Orleans, LA, USA, June 18-24, 2022}, pages
  13482--13492. {IEEE}, 2022.

\bibitem{mildenhall2020nerf}
Ben Mildenhall, Pratul~P. Srinivasan, Matthew Tancik, Jonathan~T. Barron, Ravi
  Ramamoorthi, and Ren Ng.
\newblock Nerf: Representing scenes as neural radiance fields for view
  synthesis.
\newblock In {\em ECCV}, 2020.

\bibitem{DBLP:conf/iccv/Nguyen-PhuocLTR19}
Thu Nguyen{-}Phuoc, Chuan Li, Lucas Theis, Christian Richardt, and Yong{-}Liang
  Yang.
\newblock Hologan: Unsupervised learning of 3d representations from natural
  images.
\newblock In {\em 2019 {IEEE/CVF} International Conference on Computer Vision,
  {ICCV} 2019, Seoul, Korea (South), October 27 - November 2, 2019}, pages
  7587--7596. {IEEE}, 2019.

\bibitem{DBLP:conf/nips/Nguyen-PhuocRMY20}
Thu Nguyen{-}Phuoc, Christian Richardt, Long Mai, Yong{-}Liang Yang, and
  Niloy~J. Mitra.
\newblock Blockgan: Learning 3d object-aware scene representations from
  unlabelled images.
\newblock In Hugo Larochelle, Marc'Aurelio Ranzato, Raia Hadsell,
  Maria{-}Florina Balcan, and Hsuan{-}Tien Lin, editors, {\em Advances in
  Neural Information Processing Systems 33: Annual Conference on Neural
  Information Processing Systems 2020, NeurIPS 2020, December 6-12, 2020,
  virtual}, 2020.

\bibitem{DBLP:conf/cvpr/Niemeyer021}
Michael Niemeyer and Andreas Geiger.
\newblock {GIRAFFE:} representing scenes as compositional generative neural
  feature fields.
\newblock In {\em {IEEE} Conference on Computer Vision and Pattern Recognition,
  {CVPR} 2021, virtual, June 19-25, 2021}, pages 11453--11464. Computer Vision
  Foundation / {IEEE}, 2021.

\bibitem{DBLP:conf/cvpr/ParkFSNL19}
Jeong~Joon Park, Peter Florence, Julian Straub, Richard~A. Newcombe, and Steven
  Lovegrove.
\newblock Deepsdf: Learning continuous signed distance functions for shape
  representation.
\newblock In {\em {IEEE} Conference on Computer Vision and Pattern Recognition,
  {CVPR} 2019, Long Beach, CA, USA, June 16-20, 2019}, pages 165--174. Computer
  Vision Foundation / {IEEE}, 2019.

\bibitem{Patashnik_2021_ICCV}
Or Patashnik, Zongze Wu, Eli Shechtman, Daniel Cohen-Or, and Dani Lischinski.
\newblock Styleclip: Text-driven manipulation of stylegan imagery.
\newblock In {\em Proceedings of the IEEE/CVF International Conference on
  Computer Vision (ICCV)}, pages 2085--2094, October 2021.

\bibitem{CLIP}
Alec Radford, Jong~Wook Kim, Chris Hallacy, Aditya Ramesh, Gabriel Goh,
  Sandhini Agarwal, Girish Sastry, Amanda Askell, Pamela Mishkin, Jack Clark,
  Gretchen Krueger, and Ilya Sutskever.
\newblock Learning transferable visual models from natural language
  supervision.
\newblock In Marina Meila and Tong Zhang, editors, {\em Proceedings of the 38th
  International Conference on Machine Learning, {ICML} 2021, 18-24 July 2021,
  Virtual Event}, volume 139 of {\em Proceedings of Machine Learning Research},
  pages 8748--8763. {PMLR}, 2021.

\bibitem{DBLP:conf/cvpr/RichardsonAPNAS21}
Elad Richardson, Yuval Alaluf, Or Patashnik, Yotam Nitzan, Yaniv Azar, Stav
  Shapiro, and Daniel Cohen{-}Or.
\newblock Encoding in style: {A} stylegan encoder for image-to-image
  translation.
\newblock In {\em {IEEE} Conference on Computer Vision and Pattern Recognition,
  {CVPR} 2021, virtual, June 19-25, 2021}, pages 2287--2296. Computer Vision
  Foundation / {IEEE}, 2021.

\bibitem{DBLP:conf/icip/RudinO94}
Leonid~I. Rudin and Stanley~J. Osher.
\newblock Total variation based image restoration with free local constraints.
\newblock In {\em Proceedings 1994 International Conference on Image
  Processing, Austin, Texas, USA, November 13-16, 1994}, pages 31--35. {IEEE}
  Computer Society, 1994.

\bibitem{DBLP:conf/cvpr/SanghiCLWCFM22}
Aditya Sanghi, Hang Chu, Joseph~G. Lambourne, Ye Wang, Chin{-}Yi Cheng, Marco
  Fumero, and Kamal~Rahimi Malekshan.
\newblock Clip-forge: Towards zero-shot text-to-shape generation.
\newblock In {\em {IEEE/CVF} Conference on Computer Vision and Pattern
  Recognition, {CVPR} 2022, New Orleans, LA, USA, June 18-24, 2022}, pages
  18582--18592. {IEEE}, 2022.

\bibitem{DBLP:conf/nips/SchwarzLN020}
Katja Schwarz, Yiyi Liao, Michael Niemeyer, and Andreas Geiger.
\newblock {GRAF:} generative radiance fields for 3d-aware image synthesis.
\newblock In Hugo Larochelle, Marc'Aurelio Ranzato, Raia Hadsell,
  Maria{-}Florina Balcan, and Hsuan{-}Tien Lin, editors, {\em Advances in
  Neural Information Processing Systems 33: Annual Conference on Neural
  Information Processing Systems 2020, NeurIPS 2020, December 6-12, 2020,
  virtual}, 2020.

\bibitem{DBLP:journals/corr/abs-2205-15517}
Jingxiang Sun, Xuan Wang, Yichun Shi, Lizhen Wang, Jue Wang, and Yebin Liu.
\newblock {IDE-3D:} interactive disentangled editing for high-resolution
  3d-aware portrait synthesis.
\newblock {\em CoRR}, abs/2205.15517, 2022.

\bibitem{Sun_2022_CVPR}
Jingxiang Sun, Xuan Wang, Yong Zhang, Xiaoyu Li, Qi Zhang, Yebin Liu, and Jue
  Wang.
\newblock Fenerf: Face editing in neural radiance fields.
\newblock In {\em Proceedings of the IEEE/CVF Conference on Computer Vision and
  Pattern Recognition (CVPR)}, pages 7672--7682, June 2022.

\bibitem{DBLP:conf/cvpr/SunWZLZLW22}
Jingxiang Sun, Xuan Wang, Yong Zhang, Xiaoyu Li, Qi Zhang, Yebin Liu, and Jue
  Wang.
\newblock Fenerf: Face editing in neural radiance fields.
\newblock In {\em {IEEE/CVF} Conference on Computer Vision and Pattern
  Recognition, {CVPR} 2022, New Orleans, LA, USA, June 18-24, 2022}, pages
  7662--7672. {IEEE}, 2022.

\bibitem{tov2021designing}
Omer Tov, Yuval Alaluf, Yotam Nitzan, Or Patashnik, and Daniel Cohen-Or.
\newblock Designing an encoder for stylegan image manipulation.
\newblock {\em arXiv preprint arXiv:2102.02766}, 2021.

\bibitem{DBLP:conf/cvpr/WangCH0022}
Can Wang, Menglei Chai, Mingming He, Dongdong Chen, and Jing Liao.
\newblock Clip-nerf: Text-and-image driven manipulation of neural radiance
  fields.
\newblock In {\em {IEEE/CVF} Conference on Computer Vision and Pattern
  Recognition, {CVPR} 2022, New Orleans, LA, USA, June 18-24, 2022}, pages
  3825--3834. {IEEE}, 2022.

\bibitem{DBLP:journals/corr/abs-2104-08910}
Weihao Xia, Yujiu Yang, Jing{-}Hao Xue, and Baoyuan Wu.
\newblock Towards open-world text-guided face image generation and
  manipulation.
\newblock {\em CoRR}, abs/2104.08910, 2021.

\bibitem{DBLP:journals/corr/abs-2206-07255}
Jianfeng Xiang, Jiaolong Yang, Yu Deng, and Xin Tong.
\newblock {GRAM-HD:} 3d-consistent image generation at high resolution with
  generative radiance manifolds.
\newblock {\em CoRR}, abs/2206.07255, 2022.

\bibitem{DBLP:conf/mm/YuZWZLCX0M22}
Yingchen Yu, Fangneng Zhan, Rongliang Wu, Jiahui Zhang, Shijian Lu, Miaomiao
  Cui, Xuansong Xie, Xian{-}Sheng Hua, and Chunyan Miao.
\newblock Towards counterfactual image manipulation via {CLIP}.
\newblock In Jo{\~{a}}o Magalh{\~{a}}es, Alberto~Del Bimbo, Shin'ichi Satoh,
  Nicu Sebe, Xavier Alameda{-}Pineda, Qin Jin, Vincent Oria, and Laura Toni,
  editors, {\em {MM} '22: The 30th {ACM} International Conference on
  Multimedia, Lisboa, Portugal, October 10 - 14, 2022}, pages 3637--3645.
  {ACM}, 2022.

\bibitem{DBLP:journals/corr/abs-2110-09788}
Peng Zhou, Lingxi Xie, Bingbing Ni, and Qi Tian.
\newblock {CIPS-3D:} {A} 3d-aware generator of gans based on
  conditionally-independent pixel synthesis.
\newblock {\em CoRR}, abs/2110.09788, 2021.

\bibitem{DBLP:conf/iccv/ZhuPIE17}
Jun{-}Yan Zhu, Taesung Park, Phillip Isola, and Alexei~A. Efros.
\newblock Unpaired image-to-image translation using cycle-consistent
  adversarial networks.
\newblock In {\em {IEEE} International Conference on Computer Vision, {ICCV}
  2017, Venice, Italy, October 22-29, 2017}, pages 2242--2251. {IEEE} Computer
  Society, 2017.

\bibitem{DBLP:journals/tog/ZwickerPKG02}
Matthias Zwicker, Mark Pauly, Oliver Knoll, and Markus~H. Gross.
\newblock Pointshop 3d: an interactive system for point-based surface editing.
\newblock {\em {ACM} Trans. Graph.}, 21(3):322--329, 2002.

\end{thebibliography}
}

%\title{Supplement Material}
%\input{07_Supplement.tex}

\end{document}